\documentclass[11pt]{article}
\usepackage{arxiv}

\usepackage{graphicx} 
\usepackage{amsmath, amssymb} 
\usepackage{natbib}         
\usepackage{bm}   
\usepackage{hyperref}   
\usepackage{enumitem}
\usepackage{booktabs} 
\usepackage{todonotes}
\usepackage{float}
\usepackage{algorithm} 
\usepackage{algpseudocode} 
\usepackage[bb=dsserif]{mathalpha}
\usepackage{amsthm}

\title{\texttt{findr}: Transparent and Fair Credit Risk Decisions through Semi-Structured Regressions}

\date{}

\author{ \href{https://orcid.org/0000-0002-3695-1436}{\includegraphics[scale=0.06]{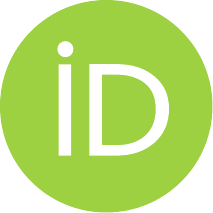}\hspace{1mm}Victor Medina-Olivares}
	\\
    University of Edinburgh, Edinburgh, UK\\
	\texttt{victor.medina@ed.ac.uk}%
    \And
	\href{https://orcid.org/0000-0001-7685-262X}{\includegraphics[scale=0.06]{assets/orcid.pdf}\hspace{1mm}Stefan Lessmann} \\
  Humboldt-Universit\"at zu Berlin, Berlin, Germany\\
    \And
    \href{https://orcid.org/0000-0003-2198-4224}{\includegraphics[scale=0.06]{assets/orcid.pdf}\hspace{1mm}Jonathan Crook}\\
    University of Edinburgh, Edinburgh, UK
}

\renewcommand{\shorttitle}{Fairness and Interpretability by Design (\texttt{findr})}
\hypersetup{
pdftitle={Fairness and Interpretability by Design (\texttt{findr})},
pdfsubject={stat.AP, q-fin.RM},
pdfauthor={Victor Medina-Olivares, Stefan Lessmann, Jonathan Crook},
pdfkeywords={Semi-structured regression, Deep Neural Networks, Fairness, Interpretability, Credit risk management},
}

\begin{document}

\maketitle

\begin{abstract}
Credit risk models increasingly need to combine predictive accuracy with transparent explanations and auditable fairness constraints. Logistic regression remains attractive because its coefficients are easy to interpret, but it can miss nonlinear structure. Flexible models can improve prediction, but their explanations are often post-hoc and may not describe the decision rule itself. We introduce \texttt{findr}, short for flexible, interpretable deep regression, a semi-structured framework for binary credit risk modelling that decomposes the logit into an interpretable structured component and an orthogonal neural residual. The orthogonalisation separates coefficient-based effects from residual nonlinear variation, while an in-processing Wasserstein penalty mitigates group disparities by comparing score distributions during training. The framework also includes diagnostics that measure the structured component's contribution to logit variation, decision agreement, and local directional consistency. We evaluate \texttt{findr} in a simulation study and on eight public credit datasets using score-level accuracy-fairness frontiers. The results show that \texttt{findr} behaves close to logistic regression when the signal is approximately linear, while recovering much of the predictive gain of neural models when nonlinear structure is relevant. The diagnostics identify when coefficient-based explanations remain close to the full fitted model and when residual variation must also be examined. These findings support semi-structured modelling as a practical way to make performance, fairness, and interpretability trade-offs explicit in credit risk decisions.
\end{abstract}

\keywords{Semi-structured regression \and Deep Neural Networks \and Fairness \and Interpretability \and Credit risk management}

\section{Introduction}
Standard logistic regression remains the workhorse of credit risk modelling, valued for its interpretability, auditability, and well-established regulatory standing \citep{dumitrescu2022machine,dastile2020statistical}. In its strictly linear log-odds specification, however, it tends to underfit the nonlinearities and feature interactions present in credit portfolios \citep{dakovic2010bankruptcy,djeundje2019identifying,medina2024deep}. Generalised additive models (GAMs) extend the same exponential-family regression logic, using additive smooth functions rather than only linear covariate effects \citep{hastie1990generalized}. In the binary case, this can be viewed as relaxing the linear predictor of logistic regression while retaining the logistic link and a degree of functional interpretability, but GAMs do not naturally capture higher-order feature interactions unless these are explicitly specified. Moreover, neither GAMs nor standard logistic regression are naturally suited to incorporating the rich unstructured data sources now available to lenders, such as transactional history, text, and images \citep{stevensonValueTextSmall2021,korangiTransformerbasedModelDefault2023}. Modern machine learning models address these representational limitations and often deliver strong predictive performance \citep{baesens2026foundation}, but at the cost of reduced transparency, since they lack inherently auditable decision logic and typically depend on post-hoc tools for interpretation \citep{gramegna2021shap}. 

This move from transparent statistical models to more flexible machine-learning systems has also changed the governance problem faced by lenders. Recent AI governance initiatives, banking supervision, and fair-lending frameworks increasingly emphasise explainability, auditability, model-risk control, and the monitoring of discriminatory or disparate impacts \citep{eu_ai_act_2024,eba2020loan,ecoa_regb}. While these sources generally do not prescribe a unique mathematical fairness criterion, they reinforce the need for credit-risk models whose predictive gains can be reconciled with transparent and accountable disparity mitigation.

In the algorithmic fairness literature, these concerns are commonly formalised through criteria that compare outcomes, errors, or score distributions across protected groups\footnote{Throughout the paper, \emph{fairness} refers to formal algorithmic criteria used to assess whether model scores, decisions, or errors differ systematically across protected groups. We distinguish this from \emph{equity}, which is a broader normative and legal concern about whether opportunities, burdens, and remedies are distributed appropriately in context. Our focus on fairness metrics is therefore an operational modelling choice. These metrics capture specific, auditable aspects of equity-relevant disparity, but they do not by themselves establish that a lending system is equitable.} \citep{hurlin2024fairness,kim2023fair}. A highly accurate model may still produce unfair outcomes for protected groups. Therefore, it is crucial to design systems that are both accurate and fair. Bias mitigation approaches are commonly grouped into \emph{pre-processing}, \emph{in-processing}, and \emph{post-processing}. Pre-processing addresses discriminatory patterns before training \citep{chen2023ai}. In-processing optimises a fairness objective during training \citep{wan2023processing}. Post-processing adjusts model outputs after training \citep{lohia2019bias}. In this work, we focus on in-processing fairness, which directly aligns the training objective with fairness goals.

There are many definitions of fairness \citep{mitchell2021algorithmic}, which can be broadly categorised into \emph{individual} and \emph{group} notions. \textbf{Individual fairness} requires that \emph{similar individuals receive similar outcomes}. It offers person-level protection and helps avoid harms that averages can hide. However, it requires a well-justified similarity metric, and defining and enforcing that metric is domain- and value-dependent and can be computationally costly at scale. \textbf{Group fairness} is often more directly auditable at the portfolio level, since firms can compare selection rates, error rates, or score distributions across protected groups. However, group-level summaries can miss within-group disparities (``fairness gerrymandering''), and several group criteria cannot generally be satisfied simultaneously when outcome prevalence differs across groups and predictions are imperfect \citep{barocas2023fairness,chouldechova2017fair}. While both paradigms have merits and limitations, we focus on \emph{group-level fairness} as our primary target, given its relevance for portfolio-level model governance and the monitoring of group-level disparities.

Following \citet{barocas2023fairness}, most group fairness criteria fall into three categories, namely \emph{independence}, \emph{separation}, and \emph{sufficiency}. \textbf{Independence} (demographic parity) asks that predictions do not differ systematically across protected groups, for example, that approval rates are similar regardless of group membership. \textbf{Separation} (error-rate parity) asks that, once the true outcome is fixed, the model behaves similarly across groups. Common special cases are \emph{equalised odds}, which matches both false-positive and true-positive rates, and \emph{equal opportunity}, which matches true-positive rates only. \textbf{Sufficiency} (calibration within groups) asks that a given risk score carries the same meaning in every group, so that applicants with a score of 0.2 default at roughly 20\% regardless of group membership. This taxonomy clarifies important tensions. When outcome prevalence differs across groups and predictions are imperfect, separation-type guarantees generally conflict with sufficiency, and independence can conflict with either. We therefore treat these criteria as complementary lenses rather than a single target. Our approach accommodates independence, equal opportunity, and equalised odds by selecting the relevant group-conditional score distributions. While sufficiency is valuable for the interpretability of risk scores, it often serves poorly as a standalone fairness target and need not be directly optimised via in-processing penalties \citep{barocas2023fairness,kozodoi2022fairness}.

We propose \texttt{findr}, a framework that integrates interpretability, flexibility, and fairness mitigation in a single end-to-end model. The name abbreviates flexible, interpretable deep regression and reflects the aim of combining deep residual flexibility with coefficient-based interpretation. The first contribution is the model itself. It decomposes the log-odds into a structured component, defined by standard interpretable features, and an unstructured component represented by a flexible representation, in our case, a neural network. Orthogonality between the two components is enforced to guarantee identifiability and transparent attribution of effects \citep{rugamer2024semi}. Fairness is encoded by penalising Wasserstein distances between the score distributions of sensitive groups, following Wasserstein fair classification \citep{jiang2020wasserstein}. A second contribution is evaluative. We introduce interpretability diagnostics that quantify the structured component's contribution to score variability and assess whether structured-component interpretations remain reliable for decisions and feature effects. We also construct score-level accuracy--fairness frontiers that provide an empirical benchmark for comparing fitted model paths at the same Wasserstein fairness level. The framework is evaluated in a simulation study and applied to eight publicly available credit-default datasets.

The following section reviews related literature. Section three describes the model, the interpretability diagnostics, and the accuracy--fairness frontier. Section four evaluates the framework using simulated data, and section five applies it to eight real obligor-level datasets. Section six concludes.

\section{Related Work}
\label{sec:related_work}

We organise the review around the two areas that motivate the framework. We first discuss interpretable credit-risk models and the question of how to explain flexible predictors. We then discuss statistical fairness criteria and penalties that reduce score-level disparities.

Credit scoring has long relied on logistic regression because it is probabilistic, auditable, and easy to communicate \citep{lessmann2015benchmarking,dastile2020statistical}. This transparency comes from a restrictive linear form for the log-odds, which can miss nonlinearities and interactions in credit portfolios \citep{dakovic2010bankruptcy,dumitrescu2022machine}. More flexible models can improve predictive fit, but their explanations often rely on tools applied after the model is fitted \citep{gramegna2021shap}. Methods based on rule extraction, distillation, and surrogate models can approximate complex models with more interpretable structures \citep{martens2007comprehensible,Tan_2023,frosst2017distilling}. Their limitation is that the explanation is not the fitted function itself. It can be informative at a global level while remaining imperfect for the local decisions that matter in lending.

Models that are interpretable by design respond to this concern by making the explanation part of the predictor itself, rather than an external approximation fitted after the fact. Generalised additive models provide readable feature effects and can relax linearity while still writing the prediction as a sum of feature effects \citep{hastie1990generalized,djeundje2019identifying}. GA2M-style models and neural additive models extend this idea through pairwise interactions or neural feature functions \citep{lou2013accurate,agarwal2021neural}. GAMI-Net and NODE-GAM further develop this line by adding structured interactions or additive architectures that can be trained by gradient methods \citep{yang2021gaminet,chang2022nodegam}. These models weaken the simple accuracy-interpretability trade-off, but their transparency still depends on restrictions on how effects and interactions enter the predictor. Semi-structured regression takes a different route by combining a conventional structured predictor with a flexible neural component \citep{rugamer2024semi}. This architecture is attractive for credit risk because the structured term can retain coefficient effects, while the neural residual captures variation that is not well represented by the structured specification. Related semi-structured ideas have been used for time to default with cure fraction and multi-state delinquency modelling \citep{medina2024deep,medina2026semi}.

The gain in flexibility creates a problem of attribution. If the structured and neural components can represent the same signal, an apparently interpretable coefficient may depend on an arbitrary allocation of fitted variation rather than on a property of the overall predictor. Orthogonalisation addresses this overlap by removing from the neural representation the part that can be explained by the structured variables \citep{rugamer2024semi}. This gives a well-defined separation between effects assigned to the coefficients and residual patterns assigned to the neural component. This differs from approaches that make additive feature effects simpler or more structured \citep{luber2023structural}. Their target is the form of the feature functions, while our target is the share of the integrated predictor that remains attributable to the structured component. Yet, identifiability is only a necessary condition for interpretation. Orthogonality does not show that the structured specification captures the relevant structure in the data, that coefficients are stable under distribution shift, or that the neural residual is small. A model can have uniquely defined coefficients while most score variation or some decisions are driven by the residual. We therefore treat interpretability as an empirical property of the fitted decomposition, assessed by how much logit variation comes from the structured component, how often structured and full-model decisions agree, and whether local effects point in the same direction.

Fairness adds a second challenge because accurate and interpretable models can still produce scores or decisions that differ systematically across protected groups. Bias-mitigation methods intervene before, during, or after model fitting \citep{hort2024bias}. Methods that act during training impose fairness through regularisation, constrained optimisation, or adversarial learning \citep{kamishima2012fairness,zafar2017fairness,agarwal2018reductions,zhang2018mitigating}. Methods applied after training can satisfy a criterion at a fixed threshold, but lending thresholds may vary across products, portfolios, economic conditions, or resource constraints \citep{de2025decision}. Training-time methods can instead act directly on the probabilistic score. This distinction matters because fairness at one decision threshold does not imply similarity of the underlying score distributions.

The choice of fairness criterion is also a subject of debate. Independence, separation, and sufficiency encode different requirements and generally cannot all hold when group base rates differ and prediction is imperfect \citep{hardt2016equality,chouldechova2017fair,barocas2023fairness}. Credit-scoring studies show that conclusions about fairness, performance, and profitability depend on both the selected criterion and the mitigation method \citep{kozodoi2022fairness,hurlin2024fairness}. Fairness assessments based on a single protected attribute can also miss intersectional disparities \citep{kim2023fair}. These findings suggest that there is no single criterion-free notion of a fair credit score. Our focus on score distributions conditional on the outcome is therefore a modelling choice, motivated by the role of error behaviour in credit-risk decisions.

Many fairness penalties that can be optimised by gradient methods compare group means or smooth approximations to threshold-based errors. These objectives are tractable, but they do not control the full distribution of scores. Two groups can have equal average predictions while differing in dispersion, tails, or the share of cases that would cross alternative decision thresholds. Fair Mixup addresses the generalisation of fairness constraints by regularising expected predictions along interpolation paths between group distributions \citep{chuang2021fair}. Wasserstein fair classification compares group score distributions more directly \citep{jiang2020wasserstein}. For univariate risk scores, the Wasserstein distance has an exact quantile representation and captures discrepancies throughout the score distribution \citep{villani2008optimal,peyre2019computational,panaretos2019statistical}.

This use of optimal transport differs from distributionally robust fairness, where robustness refers to performance under plausible shifts in the data distribution. Wasserstein distributionally robust methods define neighbourhoods around the empirical data distribution and seek models that retain performance or fairness under adverse perturbations \citep{taskesen2020distributionally,wang2020robust}. Our penalty uses Wasserstein distance for a different object. It compares protected groups within the fitted score space. The former addresses uncertainty about the population distribution, while the latter directly targets between-group disparity in model outputs.

These strands leave a gap at the intersection of fairness, flexibility, and attribution. Explanations fitted after training can describe a black box without being the decision rule. Intrinsic additive models make interpretation part of the predictor but restrict the remaining interaction structure. Semi-structured models restore flexibility and identify how fitted signal is allocated between structured and neural components, but identifiability alone does not show that the structured term explains the final score or decision. Fairness-aware classifiers can reduce selected disparities, but interventions at a fixed threshold do not control the score distribution and usually do not ask which component of a hybrid predictor absorbs the fairness adjustment.

\texttt{findr} addresses these issues jointly. It combines an identifiable structured logit with an orthogonal neural residual and applies an in-processing Wasserstein penalty to the complete probability score. Fairness is imposed on the predictor that is ultimately deployed, rather than on the structured component alone or on decisions at one threshold. At the same time, diagnostics assess whether the resulting fair score remains meaningfully attributable to its coefficient-based component. The score-level frontier then places fitted model paths on a common empirical accuracy--fairness scale. The contribution is to examine whether nonlinear fairness mitigation and structured explanation can coexist, and to make the trade-off between distributional fairness, predictive performance, and the completeness of coefficient-based attribution visible.

\section{Methodology} \label{sec:meth}

\subsection{Semi-structured logistic model}
For observation $i=1,\dots,n$, let $y_i\in\{0,1\}$ denote the default indicator ($1$ = default) and let $A_i \in\{0,1\}$ denote a sensitive attribute ($1$ = sensitive). We develop the method for this binary sensitive-attribute case, while the same construction can be extended to multi-group sensitive attributes. The vector $\bm{x}_i\in\mathbb{R}^p$ contains the covariates used in the structured linear component. The vector $\bm{z}_i$ contains the inputs to the neural network, which may coincide with $\bm{x}_i$, add further variables, or represent a separate source of information (e.g., text data). Following the semi-structured regression formulation of \citet{rugamer2024semi}, we decompose the total log-odds $\eta_i$ into a structured contribution $\eta_i^{\mathrm{str}}$ and an unstructured contribution $\eta_i^{\mathrm{unstr}}$:
\begin{equation}\label{eq:ss-logit}
\eta_i \;=\; \eta^{\mathrm{str}}_i \,+\, \eta^{\mathrm{unstr}}_i, 
\qquad 
\eta^{\mathrm{str}}_i \;=\; \beta_0 + \bm{x}_i^\top\bm{\beta}, 
\qquad 
\eta^{\mathrm{unstr}}_i \;=\; d_\theta(\bm{z}_i),
\end{equation}
where $\eta_i^{\mathrm{str}}$ is a linear logit with explicit coefficient effects, while $\eta_i^{\mathrm{unstr}}$ is a neural-network residual on the same logit scale. Here, $d_\theta$ denotes the full unstructured logit map. In the identifiable version below, this map is implemented by first applying an encoder $h_\alpha$, then orthogonalising its sample-level output, and finally applying a residual layer $r_\gamma$, with $\theta=(\alpha,\gamma)$. The default probability is $p(y_i=1\mid \eta_i)=\sigma(\eta_i)$, where $\sigma(t)=(1+\exp(-t))^{-1}$. Extensions to multiple unstructured components and generalisations to models for location, scale, and shape are also possible \citep{thielmann2024neural}.

\subsection{Identifiability via orthogonalisation}
When the same covariates enter both the structured and unstructured components in \eqref{eq:ss-logit}, the decomposition is not identifiable without additional constraints. Linear effects can then be represented either by the coefficient vector $\bm\beta$ or by the neural network, which makes attribution to the structured component ambiguous. To prevent this, we project the neural representation away from the linear span of the structured design before forming the residual logit.

Let $\bm 1_n\in\mathbb{R}^{n}$ denote the all-ones vector, and define the structured design matrix with intercept
\[
\tilde{\bm X}=[\bm 1_n\ \ \bm X]\in\mathbb{R}^{n\times(p+1)},\qquad
\bm X=(\bm x_1,\dots,\bm x_n)^\top\in\mathbb{R}^{n\times p}.
\]
With $\tilde{\bm\beta}=(\beta_0,\bm\beta^\top)^\top$, the structured logit vector is $\bm\eta^{\mathrm{str}}=\tilde{\bm X}\tilde{\bm\beta}$.
Assume $\tilde{\bm X}$ has full column rank and $n\ge p+1$. Take the thin QR decomposition
\[
\tilde{\bm X}=\bm Q \bm R,\qquad 
\bm Q\in\mathbb{R}^{n\times(p+1)}\ \text{with }\bm Q^\top \bm Q=\bm I_{p+1},\qquad
\bm R\in\mathbb{R}^{(p+1)\times(p+1)}\ \text{upper triangular}.
\]
The orthogonal projector onto the column space of $\tilde{\bm X}$ is
\[
\mathcal P=\bm Q\bm Q^\top\in\mathbb{R}^{n\times n},\qquad
\mathcal P^\perp=\bm I_n-\mathcal P\in\mathbb{R}^{n\times n}.
\]
The matrix $\mathcal P^\perp$ maps any sample-level vector onto the orthogonal complement of the column space of $\tilde{\bm X}$. Writing $\bm Z=(\bm z_1,\dots,\bm z_n)^\top$, let $\bm H_\alpha=(h_\alpha(\bm z_1),\dots,h_\alpha(\bm z_n))^\top\in\mathbb{R}^{n\times q}$ collect the neural encoder outputs before the final residual layer. Orthogonalisation is applied columnwise as
\begin{equation} \label{eq:orth}
\bm H_\alpha^\perp=\mathcal P^\perp \bm H_\alpha.
\end{equation}
Let $r_\gamma$ denote the final residual layer, with coefficients $\bm\gamma\in\mathbb{R}^q$ and intercept $\gamma_0$. In our implementation this layer is linear, so the unstructured contribution vector is
\begin{equation}\label{eq:residual-layer}
\bm\eta^{\mathrm{unstr}}
=d_\theta(\bm Z)
=r_\gamma(\bm H_\alpha^\perp)
=\bm H_\alpha^\perp\bm\gamma+\gamma_0\bm 1_n.
\end{equation}
Here $d_\theta(\bm Z)$ denotes the row-wise application of the full unstructured map to the sample. Since $\tilde{\bm X}^\top\bm H_\alpha^\perp=\bm 0$, the projected encoder features contain no intercept term and no linear combination of the covariates in $\bm X$. Apart from the residual-layer intercept $\gamma_0$, which can be absorbed into the structured intercept, linear effects are therefore attributed to the structured component, and the neural residual captures variation outside the structured linear subspace.

\subsection{In-processing fairness via Wasserstein distances}
Post-hoc fairness corrections, such as calibrating group-specific thresholds after training, can satisfy a criterion at a chosen deployment threshold but need not remain valid when thresholds change \citep{jiang2020wasserstein}. This is important in credit scoring, where thresholds shift with loan capacity, risk appetite, and regulatory requirements \citep{de2025decision}. We therefore impose the fairness penalty on the score distributions themselves. This does not guarantee a specific value of the chosen fairness criterion for every subsequent threshold, but it reduces reliance on threshold-specific post-processing by acting on the distributions from which thresholded decisions are derived.

Let $\widehat{Y}_i=\sigma(\hat{\eta}_i)\in[0,1]$ denote the predicted probability of default for observation $i$, where $\hat{\eta}_i$ is the fitted logit. Group fairness criteria compare the distributions of $\widehat{Y}$ across groups defined by $A$, optionally conditioning on the true outcome $Y$ \citep{barocas2023fairness}. 
We quantify between-group differences using the $p$-Wasserstein distance on $\mathbb{R}$, where the order $p$ is usually taken to be $1$ or $2$ \citep{shalit2017estimating}.\footnote{Our simulations and empirical analysis focus on $p=1$, but the same construction can be directly used with other Wasserstein orders.}

We first define the empirical Wasserstein distance used in the penalty. Consider two empirical probability measures on $\mathbb{R}$,
\[
\hat\mu=\sum_{i=1}^{n} \omega_i\,\delta_{s_{(i)}}\!,\qquad
\hat\nu=\sum_{j=1}^{m} \upsilon_j\,\delta_{t_{(j)}}\!,
\]
with sorted supports $s_{(1)}\le\!\cdots\!\le s_{(n)}$, $t_{(1)}\le\!\cdots\!\le t_{(m)}$ and weights $\omega\in\Delta^{n}$, $\upsilon\in\Delta^{m}$. For a ground metric $d(s,t)=|s-t|$, the discrete $p$-Wasserstein problem is
\begin{equation}\label{eq:ot-p}
W_p^p(\hat\mu,\hat\nu)
\;=\;
\min_{T\in\Pi(\omega,\upsilon)} \;\sum_{i,j} T_{ij}\,|s_{(i)}-t_{(j)}|^{p},
\qquad
\Pi(\omega,\upsilon)=\{T\!\ge\!0:\;T\mathbf{1}=\omega,\;T^\top\mathbf{1}=\upsilon\},
\end{equation}
which equals the minimum average cost of transporting $\hat\mu$ into $\hat\nu$ \citep{villani2008optimal}.

Because credit scores are one-dimensional, $W_p$ admits a closed-form characterisation via quantile matching \citep{peyre2019computational} and no entropic regularisation is required. Let $\Omega_i=\sum_{r\le i} \omega_r$ and $\Upsilon_j=\sum_{\ell\le j} \upsilon_\ell$ denote the cumulative weights of the two empirical distributions. Let $0=q_0<q_1<\cdots<q_K=1$ be the ordered grid of cumulative-mass breakpoints obtained from $\{\Omega_i\}_{i=1}^n$ and $\{\Upsilon_j\}_{j=1}^m$, with $0$ added as the left endpoint. Writing the empirical quantiles
\[
F^{-1}_{\hat\mu}(q)=s_{(i)}\ \text{for } i=\min\{r:\ \Omega_r\ge q\},\qquad
F^{-1}_{\hat\nu}(q)=t_{(j)}\ \text{for } j=\min\{\ell:\ \Upsilon_\ell\ge q\},
\]
we obtain the exact discrete formula
\begin{equation}\label{eq:wp-quantile}
W_p^p(\hat\mu,\hat\nu)
\;=\;
\sum_{k=1}^{K} (q_k-q_{k-1})\,
\big|F^{-1}_{\hat\mu}(q_{k-1})-F^{-1}_{\hat\nu}(q_{k-1})\big|^{p}.
\end{equation}
This one-dimensional form is useful for our training objective because it evaluates the fairness penalty exactly by sorting the scores and is differentiable almost everywhere, allowing gradients to be propagated through the Wasserstein term. In what follows, we write $W$ for the chosen $p$-Wasserstein distance, with $p=1$ in our empirical analysis.

We can now express the group fairness criteria from above as score-level penalties. The choice of criterion determines which group-conditional distributions of $\widehat{Y}$ are compared.
\begin{flalign}
&\textbf{Equalised Odds:}\quad
 \sum_{y\in\{0,1\}} W\!\big(\ \widehat{Y}\mid Y{=}y,A{=}0\ ,\ \widehat{Y}\mid Y{=}y,A{=}1\ \big), &&\nonumber\\[0.25em]
&\textbf{Equal Opportunity:}\quad
 W\!\big(\ \widehat{Y}\mid Y{=}0,A{=}0\ ,\ \widehat{Y}\mid Y{=}0,A{=}1\ \big), &&\nonumber\\[0.25em]
&\textbf{Independence:}\quad
 W\!\big(\ \widehat{Y}\mid A{=}0\ ,\ \widehat{Y}\mid A{=}1\ \big). &&\label{eq:criteria}
\end{flalign}

Equalised odds compares score distributions separately within each realised outcome group, so disparities are penalised among both non-defaulters and defaulters. Equal opportunity focuses only on the outcome group treated as opportunity-relevant in this application, here $Y=0$, and therefore targets score disparities among non-defaulters. Independence does not condition on the outcome and instead compares the overall score distributions across protected groups.

For a chosen criterion, we set $\mathcal L_\text{fair}(\widehat{Y},A,Y)$ to the corresponding scalar penalty in \eqref{eq:criteria}, computed via \eqref{eq:wp-quantile}.

\subsection{Training objective}
\begin{figure}
  \centering
  \includegraphics[width=0.9\linewidth]{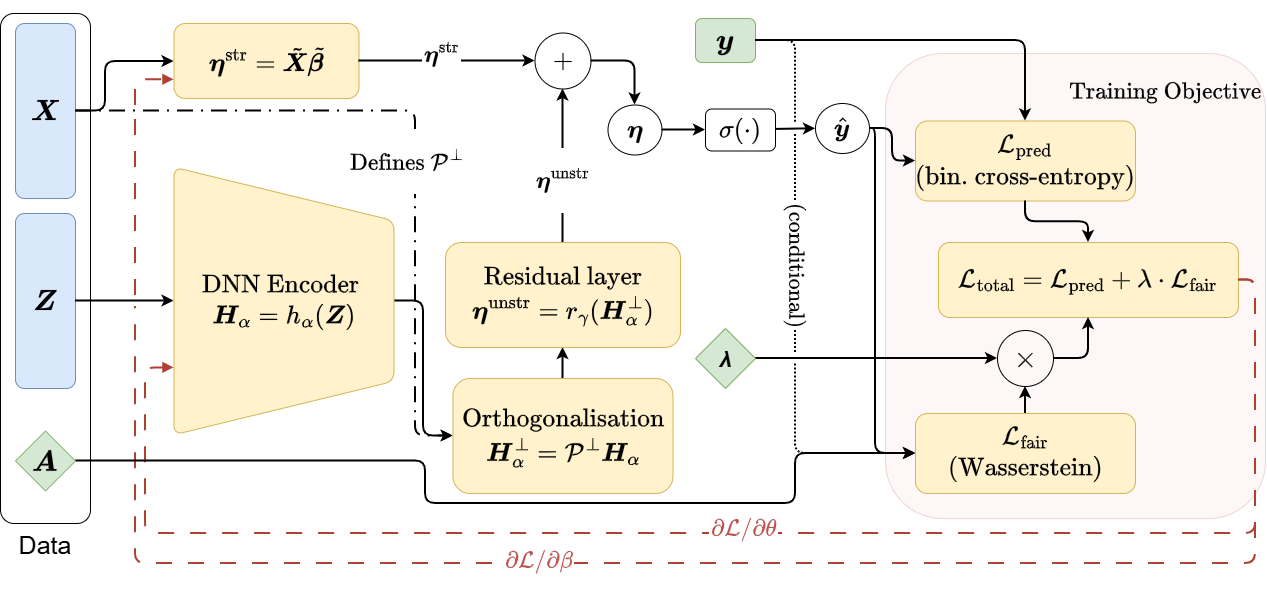}
  \caption{\texttt{findr} architecture.}
  \label{fig:findr-architecture}
\end{figure}

The total loss is a weighted combination of predictive fit and a fairness penalty,
\begin{equation}\label{eq:total-loss}
\mathcal{L}_\text{total}(\theta,\beta) \;=\; 
\underbrace{\frac{1}{n}\sum_{i=1}^n \big[-y_i\log \widehat{Y}_i - (1-y_i)\log(1-\widehat{Y}_i)\big]}_{\mathcal{L}_\text{pred}}
\;+\; \lambda \cdot \mathcal{L}_\text{fair}\big(\widehat{Y},A,Y\big),
\end{equation}
with $\lambda\!\ge\!0$ governing the accuracy-fairness trade-off. Gradients are backpropagated through both the orthogonalisation step and the Wasserstein distance to update $(\theta,\beta)$ via stochastic optimisation. Figure~\ref{fig:findr-architecture} summarises the resulting architecture.

\subsection{Score-level accuracy--fairness frontier}
\label{sec:frontier}
The training objective above produces a family of fitted models indexed by the fairness multiplier $\lambda$. To evaluate these models, we compare them with a score-fairness reference frontier. The frontier asks how much predictive accuracy is attainable at a given value of the same Wasserstein unfairness criterion used for training. This differs from standard accuracy--fairness frontiers, which are usually defined for thresholded decisions through false-positive and false-negative rates \citep{hardt2016equality,menon2018cost}. Our penalty is applied before thresholding, so the fairness axis compares distributions of probability scores.

Let $\mathcal{D}_{\mathrm{val}}=\{(x_i,y_i,a_i)\}_{i=1}^n$ be the validation sample. Let $r_i\in(0,1)$ denote a reference estimate of $P(Y=1\mid X=x_i)$, using the true conditional risk in simulations when available and otherwise the score from an unconstrained model fitted on the training data. A candidate predictor assigns one score $s_i\in(0,1)$ to each validation observation. This mirrors how fitted models are evaluated, since each individual receives one probability score, and the realised labels and groups are then used only to compute prediction loss and fairness gaps.

For equalised odds, the empirical unfairness of a candidate score is
\begin{equation}
    \widehat{\Phi}(s;y,a)
    = \sum_{y\in\{0,1\}} W\!\left(\widehat\mu_{y0}^s,\widehat\mu_{y1}^s\right),
    \label{eq:phi}
\end{equation}
where $\widehat\mu_{ya}^s$ is the empirical distribution of $s_i$ among validation observations with $Y_i=y$ and $A_i=a$. For $p=1$, this equals the integral of the absolute difference between the two empirical CDFs over all score cutoffs. Thus $y_i$ and $a_i$ enter the frontier through the fairness functional, because equalised odds is defined by distributions conditional on realised outcome and sensitive group. We focus on equalised odds here, but the same procedure applies to the other criteria in \eqref{eq:criteria} by replacing $\widehat\Phi$ with the corresponding empirical score-distribution penalty.

Predictive accuracy is measured by the expected Bernoulli log loss of $s$ under the reference risk $r$,
\begin{equation}
    \widehat R_{\log}(r,s)
    = -\frac{1}{n}\sum_{i=1}^n
      \bigl[r_i\log s_i+(1-r_i)\log(1-s_i)\bigr]
    = \widehat H(r)+\frac{1}{n}\sum_{i=1}^n \mathrm{KL}\{\mathrm{Bern}(r_i)\,\|\,\mathrm{Bern}(s_i)\}.
    \label{eq:rlog}
\end{equation}
Here $\widehat H(r)$ is the empirical Bernoulli entropy of the reference risks, which is the expected log loss obtained by predicting $r_i$ itself. The second equality therefore writes the expected log loss as this irreducible term plus the average KL divergence from $r$ to $s$. This expression does not use the realised $y_i$ because it is the conditional expectation of binary log loss given $X_i$ under $P(Y=1\mid X_i)=r_i$. It therefore measures the expected loss relative to the reference risk surface rather than sampling noise in a particular validation draw. The unconstrained optimum is $s=r$, and plots report the excess log loss $\widehat R_{\log}(r,s)-\widehat H(r)$ so that this point has value zero.

The predictor frontier is the lower boundary of
\[
    \min_s \widehat R_{\log}(r,s)
    \quad \text{subject to} \quad
    \widehat\Phi(s;y,a)\le \tau,
\]
as $\tau$ varies. Numerically, we trace this boundary by solving the scalarised problem
\begin{equation}
    \min_s\; Q_\xi(s)
    = \widehat R_{\log}(r,s)+\xi\,\widehat\Phi(s;y,a),
    \qquad \xi\ge0,
    \label{eq:frontier-scalarisation}
\end{equation}
over a grid of frontier multipliers $\xi$. We use $\xi$ rather than $\lambda$ to separate the post-hoc frontier construction from the model-training multiplier in \eqref{eq:total-loss}. At $\xi=0$, the solution is $s=r$. Larger values of $\xi$ move the score vector toward lower empirical unfairness, usually at the cost of higher expected log loss. The resulting points, together with any candidate model scores, are sorted by $\widehat\Phi$ and converted to a monotone lower envelope. Appendix~\ref{app:predictor-frontier} gives the computational details. The frontier should be interpreted as a numerical score-vector benchmark rather than as the attainable boundary for any specified family of fitted models. It gives the best empirical trade-off found by the construction on the validation sample, relative to the reference risk $r$, and is therefore useful for comparing fitted models on a common scale rather than for establishing a population-optimal accuracy--fairness bound.

For a fitted model with validation scores $s_m$, the accuracy gap is
\begin{equation}
    \Delta(s_m)
    = \widehat R_{\log}(r,s_m)-F\!\left(\widehat\Phi(s_m;y,a)\right),
    \label{eq:gap}
\end{equation}
where $F$ is the interpolated frontier value. A smaller gap means that the fitted model is closer to the best score vector found at the same fairness level. The quantity is empirical and depends on the numerical frontier construction, so small negative values can occur if the computed envelope fails to dominate an unseeded candidate score.

\subsection{Diagnostics for interpretable-component contribution}
Beyond coefficient readability, we use three diagnostics to quantify the contribution of the interpretable structured component to the fitted model. These diagnostics assess its role in score variation, binary decisions, and feature-level directional interpretations. Throughout this subsection, $\bm\eta$ denotes the full logit vector, $\bm\eta^{\mathrm{str}}$ denotes the structured logit vector, and $\bm\eta^{\mathrm{unstr}}$ denotes the neural residual from \eqref{eq:ss-logit}.

\subsubsection{Explained Variability Ratio (EVR)}
EVR measures the share of logit-scale variation attributable to the structured component. We define
\begin{equation}\label{eq:evr}
\mathrm{EVR}
\;=\; \frac{\mathrm{Var}_n\!\big(\bm{\eta}^{\mathrm{str}}\big)}{\mathrm{Var}_n\!\big(\bm{\eta}\big)}
\;=\;
\frac{\frac{1}{n}\sum_i \big(\eta^{\mathrm{str}}_i-\bar\eta^{\mathrm{str}}\big)^2}
  {\frac{1}{n}\sum_i \big(\eta_i-\bar\eta\big)^2}.
\end{equation}
The orthogonalisation in \eqref{eq:orth} makes the total logit variance decompose into the sum of structured and residual variation. Hence
\[
\mathrm{EVR}=\frac{\mathrm{Var}_n(\bm{\eta}^{\mathrm{str}})}{\mathrm{Var}_n(\bm{\eta}^{\mathrm{str}})+\mathrm{Var}_n(\bm{\eta}^{\mathrm{unstr}})}.
\]
Thus EVR can be interpreted as the structured component's share of total logit-scale variation. A high EVR indicates that most score variation is explained by the transparent linear component. A low EVR indicates that the neural residual accounts for a material share of the fitted score variation.

\subsubsection{Decision Disagreement Rate (DDR)}
EVR is measured on the continuous logit scale. DDR complements it by measuring whether the structured component and the full model imply the same binary decision at the default threshold. Let
\[
\hat y_i \;=\; \mathbb{1} [\eta_i \ge 0],
\qquad
\hat y_i^{\mathrm{str}} \;=\; \mathbb{1} [\eta_i^{\mathrm{str}} \ge 0]
\]
denote the class labels implied by the full and structured logits. The decision disagreement rate is
\begin{equation}\label{eq:ddr}
\mathrm{DDR}
\;=\;
\frac{1}{n}\sum_{i=1}^n
\mathbb{1}\!\left(\hat y_i^{\mathrm{str}} \neq \hat y_i\right).
\end{equation}
DDR is zero when the structured component and the full model make identical classifications on the evaluation sample. Larger values indicate that using the structured component alone would often lead to different decisions than using the integrated model.

\subsubsection{Counterfactual Sign Error Rate (CSER)}
The structured coefficients also provide feature-level directional interpretations. CSER checks whether these directions remain consistent after adding the neural residual. For each structured feature $j$ with $\beta_j\neq0$, let $d_j=\mathrm{sign}(\beta_j)$ and let $\delta>0$ denote a prespecified perturbation size. Define
\[
\bm{x}_i^{(j,+)} = \bm{x}_i + d_j\,\delta\,\bm{e}_j,
\]
where $\bm{e}_j$ is the $j$th unit vector. This perturbation moves feature $j$ in the direction that the structured coefficient associates with higher default log-odds. The counterfactual sign error rate for feature $j$ is
\begin{equation}\label{eq:cser}
\mathrm{CSER}_j
\;=\;
\frac{1}{n}
\sum_{i=1}^n
\mathbf{1}_{\{\,
\eta(\bm{x}_i^{(j,+)},\bm{z}_i)
<
\eta(\bm{x}_i,\bm{z}_i)
\,\}}.
\end{equation}
CSER$_j$ is the share of observations for which a risk-increasing perturbation according to the structured coefficient lowers the full model logit. It therefore measures directional contradictions between the transparent coefficient and the integrated model. In the tables, we report the minimum and maximum CSER across structured features,
\[
\mathrm{CSER}_{\min}=\min_{j:\beta_j\neq0}\mathrm{CSER}_j,
\qquad
\mathrm{CSER}_{\max}=\max_{j:\beta_j\neq0}\mathrm{CSER}_j,
\]
to summarise the range of feature-level sign reliability.

\section{Simulation Study}
The simulation is designed to study three questions. First, we ask whether \texttt{findr} behaves like a transparent linear model when the true signal is linear. Second, we ask whether the neural residual recovers additional predictive structure when the signal is nonlinear. Third, we analyse how the Wasserstein penalty moves fitted scores along the accuracy--fairness trade-off while preserving the contribution of the structured component.
\subsection{Data-generating processes}
We simulate $n=10{,}000$ observations with $p=6$ covariates. The sensitive attribute and pre-shift covariates follow
\[
A\sim\mathrm{Bern}(0.3), \qquad X_j\sim\mathcal{N}(0,1)\;\;(j=1,\dots,6).
\]
We induce two sources of score-distribution unfairness. The first is a group-dependent covariate shift, $X_1 \leftarrow X_1 + (1-A)$. The second is a direct group-specific logit shift, $B(A)=0.2\,\bm{1}\{A=1\}$. The linear component is $L=b_0+\bm{\beta}^\top X$, with $b_0=-2$ and $\bm\beta=(-1,-0.5,0,0.25,0.5,1)^\top$. To create nonlinear structure, we add univariate nonlinear terms $\mathrm{NL}(X)=\sum_{k\in\mathcal{I}_{\mathrm{nl}}} g_k(X_k)$ and interactions $\mathrm{I}(X)=\sum_{(i,j)\in \mathcal{I}_{\mathrm{int}}} X_iX_j$. We consider two data-generating processes:
\begin{itemize}
  \item \textbf{Simple mode}. $\eta=L+B(A)$, so the true logit is linear with both covariate-shift and baseline unfairness.
  \item \textbf{Complex mode}. $\eta=L+\mathrm{NL}(X)+\mathrm{I}(X)+B(A)$, so the true logit contains linear, nonlinear, and interaction effects, again with both sources of unfairness.
\end{itemize}
Outcomes are drawn from $Y\sim\mathrm{Bern}(\sigma(\eta))$. The simple mode therefore tests whether the semi-structured model avoids using the residual when it is unnecessary. The complex mode tests whether the same architecture can recover nonlinear signal without losing the linear component used for interpretation.

\subsection{Models and training}
We compare three models. \texttt{LR} is a standard logistic regression and represents the transparent linear baseline. \texttt{NN} is a two-layer feed-forward neural logistic model with the same neural backbone as \texttt{findr}, but without a separate structured component. \texttt{findr} is the semi-structured model in Section~\ref{sec:meth}, with a linear structured logit and an orthogonal neural residual.

All models are trained with the objective in \eqref{eq:total-loss}. We fit the three fairness penalties described in \eqref{eq:criteria} in turn. We vary the multiplier $\lambda$ over a grid from $0$ to $4$ and repeat each configuration over $10$ random seeds. Some visual summaries display only the range up to $\lambda=1$ when larger penalties do not add substantive information. We report the equalised-odds results because this criterion conditions on both outcome classes, making it the most demanding of the three criteria considered here. The score-level frontier in Section~\ref{sec:frontier} is constructed for the same criterion, and the other criteria led to the same qualitative comparisons across models.

\subsection{Evaluation}
We evaluate the simulations along the same three dimensions used in the rest of the paper. Predictive performance is measured out of sample using AUC and Brier score. Fairness is measured by the equalised-odds Wasserstein penalty, and the frontier places model paths on the common score-level accuracy--fairness scale developed in Section~\ref{sec:frontier}. Interpretable-component contribution is assessed using the coefficient paths, EVR, DDR, and CSER from the diagnostics in Section~\ref{sec:meth}.

\begin{figure}[!t]
  \centering
  \includegraphics[width=0.9\linewidth]{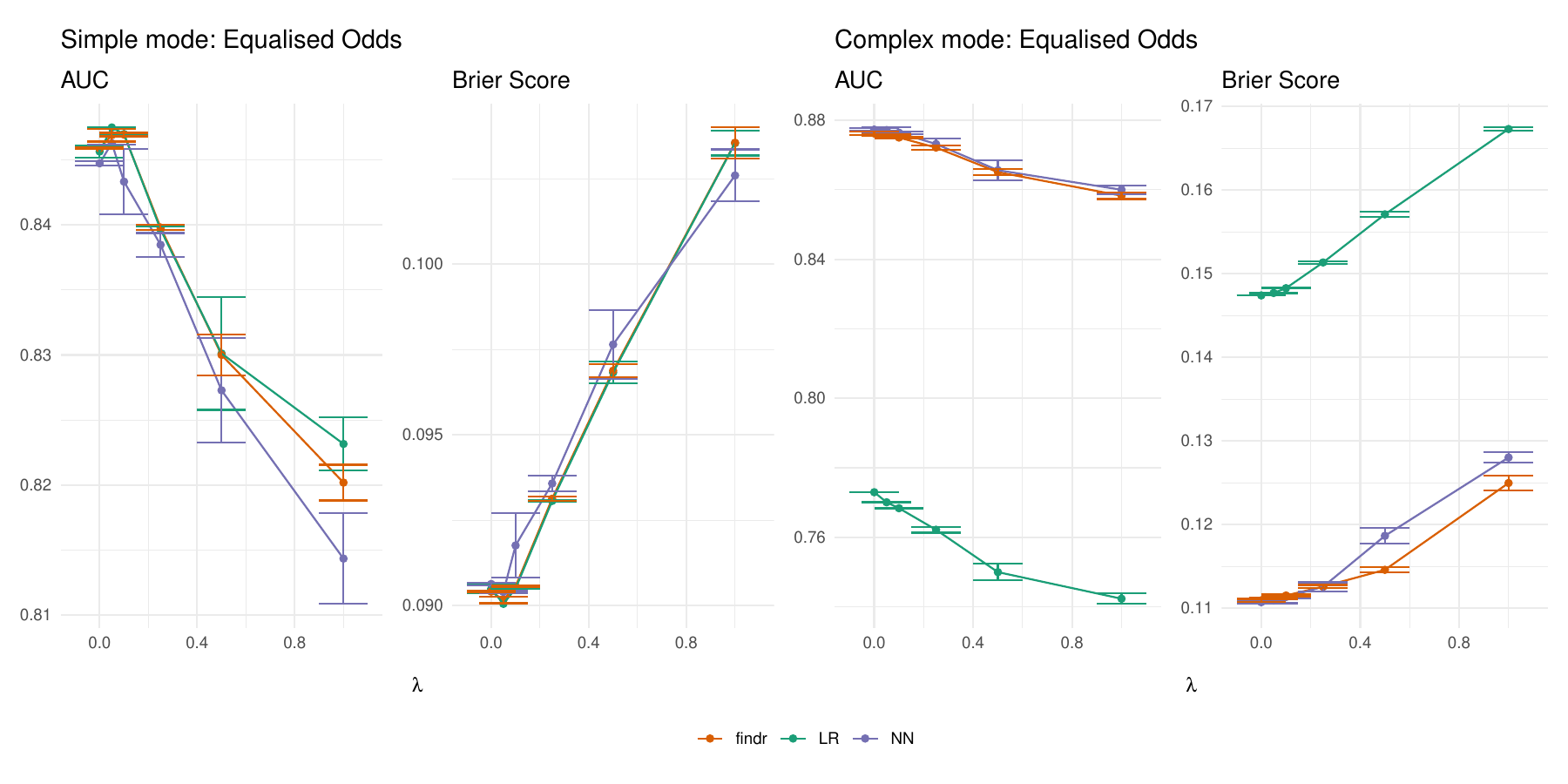}
  \caption{Out-of-sample performance in the simple and complex simulation settings. Curves show the mean across $10$ random seeds as the fairness multiplier $\lambda$ varies under equalised odds.}
  \label{fig:sim-performance}
\end{figure}

\paragraph{Out-of-sample performance.}
Figure~\ref{fig:sim-performance} compares predictive performance as $\lambda$ varies. In the simple mode, the true logit is linear, so logistic regression is correctly specified. The neural model and \texttt{findr} therefore have little room to improve predictive accuracy over \texttt{LR}. The relevant check is whether \texttt{findr} preserves this linear behaviour rather than introducing unnecessary nonlinear variation. In the complex mode, the neural residual has a role because the data-generating logit includes nonlinear terms and interactions. The expected pattern is therefore different, since \texttt{LR} is misspecified, while \texttt{NN} and \texttt{findr} can use the neural component to recover additional predictive signal.

In the simple mode, \texttt{findr} behaves like the linear model because the orthogonal residual has little unexplained structure to capture. In the complex mode, \texttt{findr} instead follows the neural model more closely than the linear baseline, indicating that the residual component captures nonlinear signal while the structured component remains available for interpretation. As $\lambda$ increases, performance can decline because the fitted scores move toward smaller score-distribution gaps.

\begin{figure}[!t]
  \centering
  \includegraphics[width=0.9\linewidth]{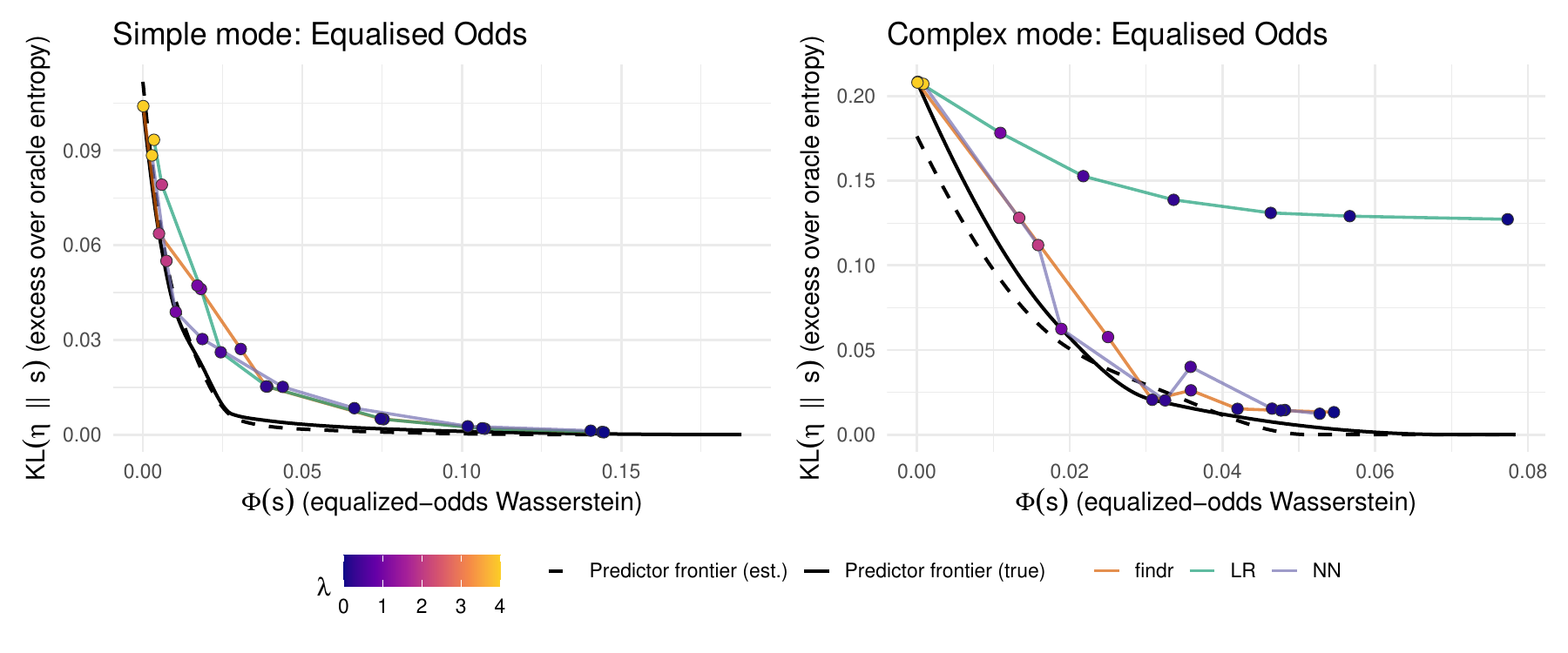}
  \caption{
Predictor frontiers and model paths under equalised odds.
The panels compare the simple and complex simulation settings.
Black curves show the true and estimated predictor frontiers in the fairness--log-loss plane, where the horizontal axis is the equalised-odds Wasserstein distance $\widehat\Phi(s;y,a)$ and the vertical axis is the excess log loss relative to the reference-risk entropy $\widehat H(r)$.
Coloured lines trace the solutions obtained by each model as the regularisation parameter $\lambda$ varies, and the point colour gradient indicates the value of $\lambda$.
Points closer to the lower-left region correspond to lower fairness violation and lower predictive loss.
}
  \label{fig:sim-frontier}
\end{figure}

\paragraph{Fairness.}
Figure~\ref{fig:sim-frontier} places the fitted model paths against the predictor frontier. The horizontal axis is the empirical equalised-odds Wasserstein distance $\widehat\Phi(s;y,a)$, and the vertical axis is the excess expected log loss relative to the reference-risk entropy $\widehat H(r)$. Points nearer the lower-left region have both smaller fairness violation and smaller prediction loss. The black frontier is a score-vector benchmark rather than a model-class boundary, so the comparison asks how close each fitted model comes to the best score trade-off found on the validation sample.

The model paths illustrate how the in-processing penalty acts on the full score distribution. At $\lambda=0$, models prioritise predictive fit and can retain the group differences induced by the covariate and baseline shifts. As $\lambda$ increases, their scores move toward lower equalised-odds distance. The trade-off is mild when the fairness correction is aligned with the available model structure, and stronger when fairness constraints require score changes that also reduce predictive fit.

\begin{figure}[!t]
  \centering
  \includegraphics[width=0.9\linewidth]{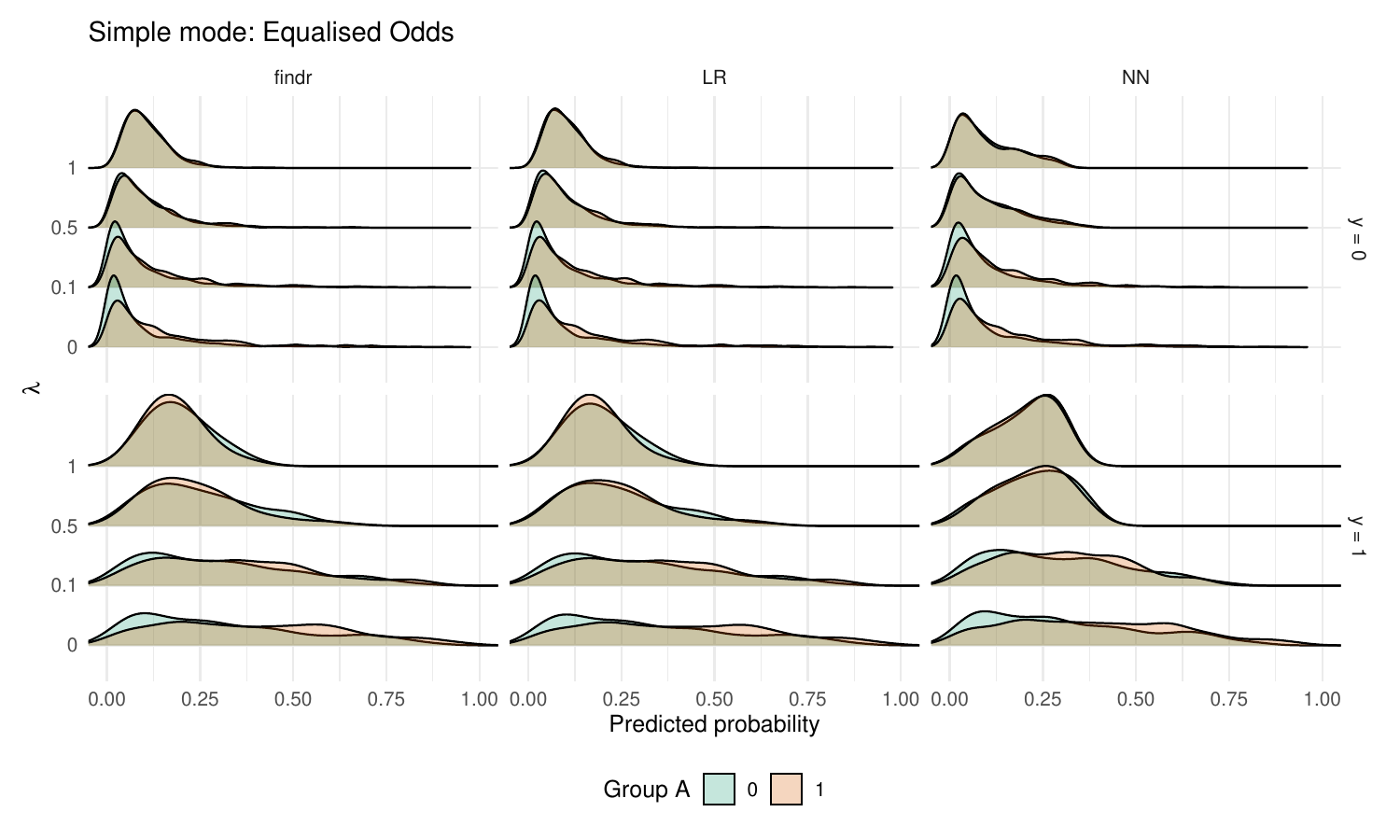}
  \caption{Score distributions in the simple simulation setting. The panels show fitted score distributions across groups and outcome strata as $\lambda$ varies.}
  \label{fig:sim-density-simple}
\end{figure}

\begin{figure}[!t]
  \centering
  \includegraphics[width=0.9\linewidth]{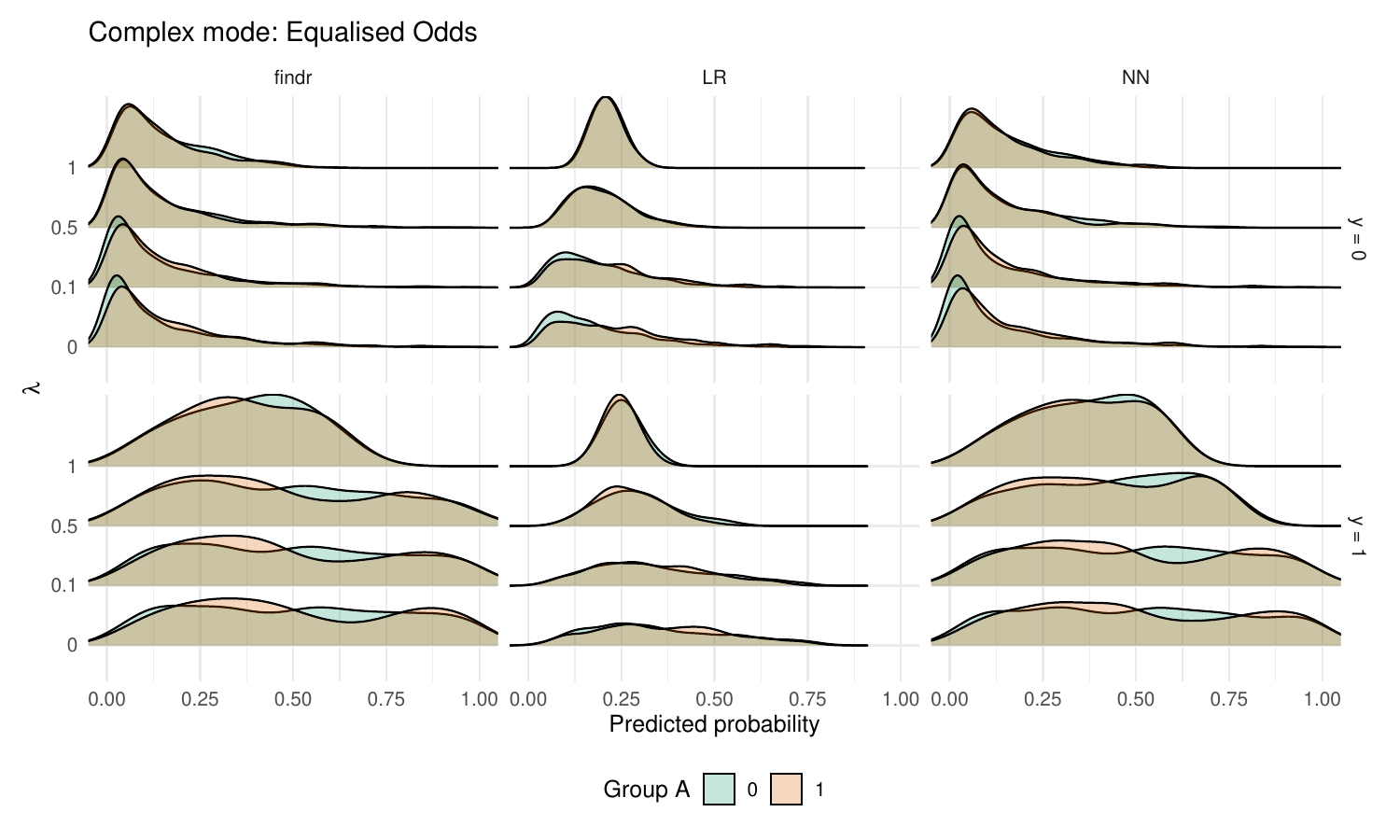}
  \caption{Score distributions in the complex simulation setting. The panels show fitted score distributions across groups and outcome strata as $\lambda$ varies.}
  \label{fig:sim-density-complex}
\end{figure}

Figures~\ref{fig:sim-density-simple} and~\ref{fig:sim-density-complex} show the same mechanism at the distributional level. As expected, the penalty does not adjust a single threshold but changes the fitted score distributions within the outcome-by-group cells that define equalised odds. As $\lambda$ increases, these group-conditional distributions move closer within each model. This does not require the final score distributions to have the same shape across model specifications. Different model classes can satisfy a smaller Wasserstein gap while still producing distinct score distributions. This agrees with the motivation for using Wasserstein penalties in Section~\ref{sec:meth}, where the fairness intervention is applied before thresholding.
 
\begin{figure}[!t]
  \centering
    \includegraphics[width=0.9\linewidth]{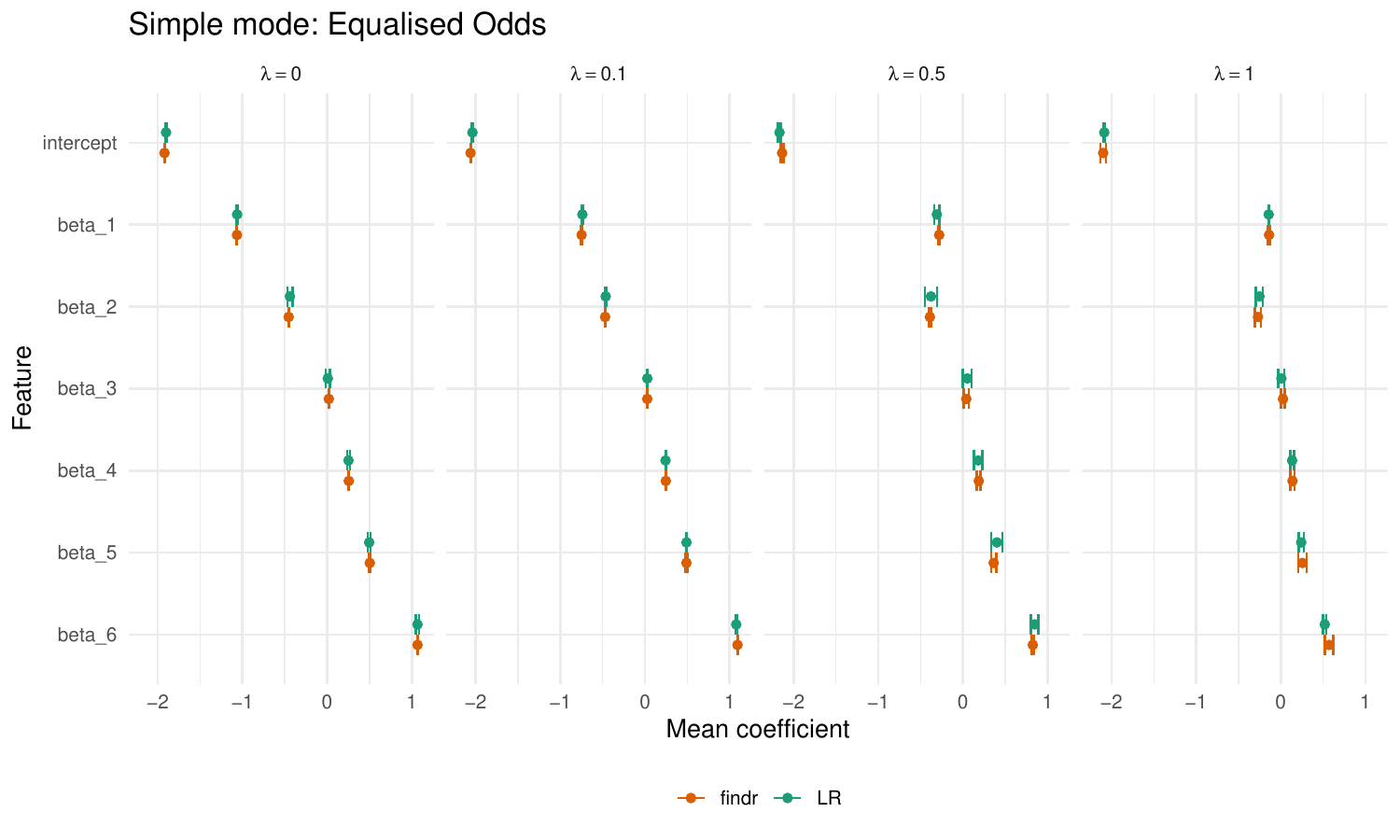}
  \caption{Estimated structured coefficients in the simple simulation setting. The true signal is linear, and the structured coefficients from \texttt{findr} follow the same coefficient pattern as \texttt{LR}.}
  \label{fig:sim-betas-simple}
\end{figure}

\paragraph{Interpretability.}
We next study how much of the fitted model can be assigned to the structured component. In the simple mode, the true signal is linear, so \texttt{findr} should behave like a linear model with little need for the residual. The coefficient paths in Figure~\ref{fig:sim-betas-simple} follow the same pattern as \texttt{LR}, including the signs of the linear effects. The diagnostics in Table~\ref{tab:tab:sim-evr-ddr-csermin-csermax-metrics-simple} lead to the same conclusion. EVR remains at or near one across the reported $\lambda$ values, DDR is essentially zero, and CSER is negligible. Thus, the full fitted logit is almost entirely explained by the structured component, and using the structured component alone leads to the same decisions for nearly all observations. This is the desired result in a setting where the residual has no genuine nonlinear signal to recover.

Figure~\ref{fig:sim-betas-simple} also shows how the fairness penalty affects the structured component. Recall that the simulation setup encodes feature-level disparities in covariate $X_1$. Accordingly, increasing $\lambda$ shrinks the effect of that feature, $\beta_1$, toward zero, whereas the effects of other covariates change relatively little.  

\begin{table}[!h]
\centering
\caption{\label{tab:tab:sim-evr-ddr-csermin-csermax-metrics-simple}\textbf{Simple mode:} Equalised Odds. EVR, DDR, CSER (min) and CSER (max) reported as mean (SD) for the findr model by $\lambda$.}
\centering
\begin{tabular}[t]{ccccc}
\toprule
\(\lambda\) & EVR & DDR & CSER (min) & CSER (max)\\
\midrule
0.0 & 1.000 (0.000) & 0.000 (0.000) & 0.000 (0.000) & 0.000 (0.000)\\
0.1 & 1.000 (0.000) & 0.001 (0.001) & 0.000 (0.000) & 0.000 (0.000)\\
0.5 & 0.997 (0.009) & 0.002 (0.004) & 0.000 (0.000) & 0.002 (0.006)\\
1.0 & 0.984 (0.019) & 0.001 (0.001) & 0.000 (0.000) & 0.005 (0.014)\\
\bottomrule
\end{tabular}
\end{table}

\begin{figure}[!t]
  \centering
    \includegraphics[width=0.9\linewidth]{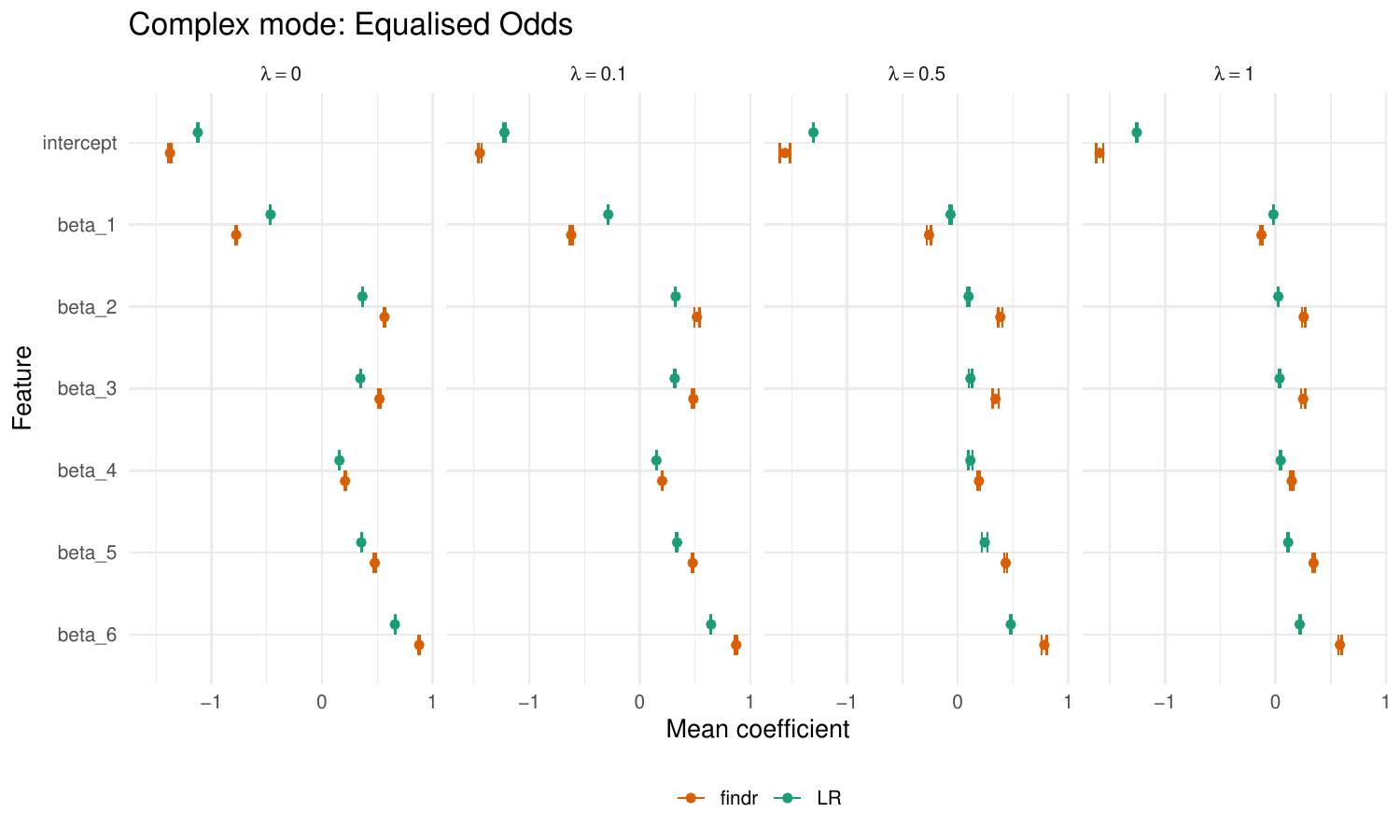}
  \caption{Estimated structured coefficients in the complex simulation setting. The structured coefficients from \texttt{findr} remain coherent with \texttt{LR}, while the neural residual accounts for nonlinear terms and interactions.}
  \label{fig:sim-betas-complex}
\end{figure}

In the complex mode, the signal is split by design between linear terms, nonlinear terms, and interactions. The residual should therefore carry a larger share of fitted score variation. The relevant question is not whether all signal remains structured, but whether \texttt{findr} can still identify the part that is structured and interpretable. The coefficient paths in Figure~\ref{fig:sim-betas-complex} remain coherent with \texttt{LR}, which indicates that the structured component still captures the linear part of the signal. At the same time, Table~\ref{tab:tab:sim-evr-ddr-csermin-csermax-metrics-complex} shows that EVR is lower than in the simple mode, ranging from about $0.44$ to $0.56$ across the reported $\lambda$ values. This means that a substantial share of logit-scale variation is assigned to the neural residual rather than to the structured component. DDR also becomes nonzero, so some decisions depend on the integrated model rather than the structured component alone. CSER remains small for some features but can increase for the most affected feature as $\lambda$ grows, indicating that fairness regularisation and nonlinear residual structure can alter local directional behaviour. This is also shown in Figure~\ref{fig:sim-betas-complex}, where the effect of increases in $\lambda$ is particularly pronounced for $\beta_1$. Together, these results show that \texttt{findr} retains coefficient interpretations coherent with the logistic baseline while also quantifying how much fitted variation is carried by the structured component and how much is carried by the residual. A standard logistic model cannot make this split because it has no residual component, and a pure neural network cannot make it because it has no separate interpretable component.

\begin{table}[!h]
\centering
\caption{\label{tab:tab:sim-evr-ddr-csermin-csermax-metrics-complex}\textbf{Complex mode:} Equalised Odds. EVR, DDR, CSER (min) and CSER (max) reported as mean (SD) for the findr model by $\lambda$.}
\centering
\begin{tabular}[t]{ccccc}
\toprule
\(\lambda\) & EVR & DDR & CSER (min) & CSER (max)\\
\midrule
0.0 & 0.556 (0.013) & 0.123 (0.004) & 0.000 (0.000) & 0.038 (0.007)\\
0.1 & 0.525 (0.006) & 0.120 (0.001) & 0.000 (0.000) & 0.055 (0.013)\\
0.5 & 0.436 (0.016) & 0.112 (0.004) & 0.000 (0.000) & 0.289 (0.032)\\
1.0 & 0.469 (0.017) & 0.068 (0.003) & 0.011 (0.006) & 0.424 (0.020)\\
\bottomrule
\end{tabular}
\end{table}

\section{Application} \label{sec:app}
To complement the results of the simulation study, we proceed with comparing \texttt{findr} and its two benchmarks across real-world lending data.
\subsection{Datasets and sensitive attribute}
We study eight public datasets frequently used in credit-scoring modelling. All datasets use \emph{age} as the sensitive attribute (denoted as $A$ in Section~\ref{sec:meth}), binarised at $25$ (``younger'' vs.\ ``older''), which has been a common setting in previous work \citep{kozodoi2022fairness, kim2023fair, kamiranClassifyingDiscriminating2009}. Below, we report the sample size $n$, number of features, default rate, and sensitive-class share (proportion with age $\le 25$) for each dataset, with further details in Appendix~\ref{app:data}.

\begin{description}
    \item[South German Credit.] $n=1{,}000$ applicants with $19$ features; default rate $30.0\%$; sensitive class share $19.0\%$.
    \item[Taiwan Credit Default.] $n=30{,}000$ clients with $22$ features; default rate $22.12\%$; sensitive class share $12.9\%$.
    \item[Give Me Some Credit.] $n=150{,}000$ applicants with $9$ features; default rate $6.68\%$; sensitive class share $2.02\%$.
    \item[Vehicle Car Loan.] $n=233{,}154$ loans with $32$ features; default rate $21.71\%$; sensitive class share $22.84\%$.
    \item[MyHome Loan.] $n=7{,}000$ loans with $7$ features; default rate $40.0\%$; sensitive class share $7.03\%$.
    \item[Thomas Loan.] $n=1{,}225$ applicants with $13$ features; default rate $26.37\%$; sensitive class share $7.67\%$.
    \item[UK Credit.] $n=30{,}000$ applicants with $13$ features; default rate $4.0\%$; sensitive class share $19.75\%$.
    \item[PAKDD 2010.] $n=50{,}000$ applicants with $51$ features; default rate $26.08\%$; sensitive class share $11.49\%$.
\end{description}

We split the data into training, validation, and test sets using a 70/10/20 split. Continuous features are standardised, and categorical variables are encoded using weights of evidence. We train \texttt{LR}, \texttt{NN}, and \texttt{findr} across a $\lambda$ grid ranging from 0 to 2 and repeat each configuration over 10 random seeds. Hyperparameters are selected on the validation split, with details given in Appendix~\ref{app:hyperparameter-search}, and all reported metrics are evaluated out of sample on the test split. As in the simulation study, the main text reports equalised-odds results. The other fairness criteria were also examined, but they did not add new qualitative insights to the model comparisons.

\subsection{Results}

\subsubsection{Out-of-sample performance} \label{subsec:app_perf}

\begin{figure}[!t]
  \centering
  \includegraphics[width=0.9\linewidth]{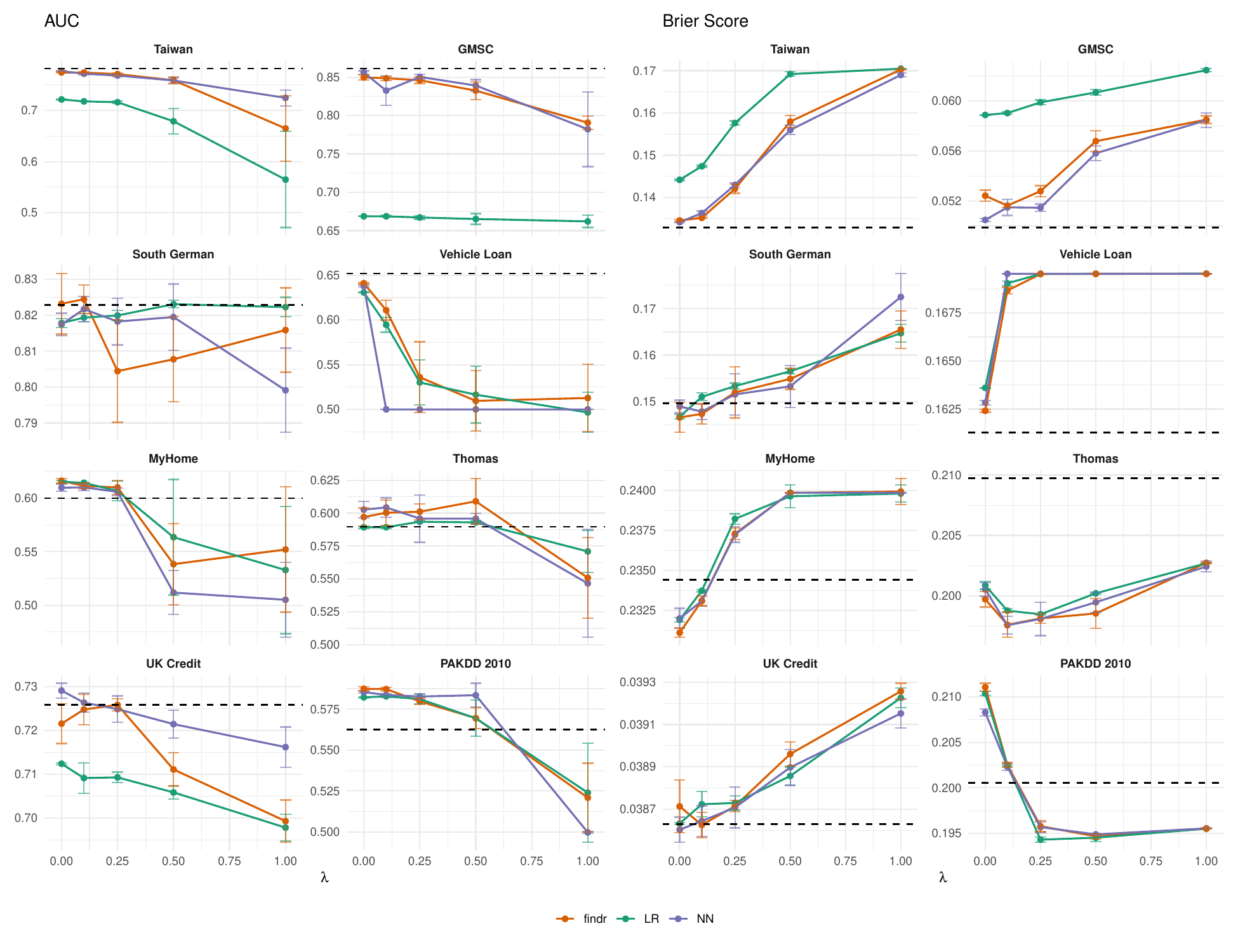}
  \caption{Out-of-sample discrimination and calibration across datasets under the equalised-odds penalty. Curves show AUC and Brier score as the fairness multiplier $\lambda$ varies. Dashed horizontal lines show the unpenalised CatBoost reference.}
  \label{fig:app-performance}
\end{figure}

Figure~\ref{fig:app-performance} summarises out-of-sample discrimination and calibration across the $\lambda$ path, measured by AUC and Brier score. The dashed horizontal line gives the unpenalised CatBoost result \citep{prokhorenkova2018catboost}, included as a state-of-the-art tabular-data reference that is neither fairness-constrained nor interpretable. It provides a scale for assessing how close the three penalised model classes remain to a strong predictive benchmark as $\lambda$ varies. 

The plotted paths show that the predictive effect of the fairness penalty is not uniform across datasets. In some datasets, AUC and Brier score remain relatively stable along the path, while in others stronger penalisation moves the fitted scores toward weaker discrimination or poorer calibration.

Tables~\ref{tab:summary_lam0} and~\ref{tab:summary_lam01} give two cuts of these paths, at $\lambda=0$ and $\lambda=0.1$, to make the contrast between empirical predictive regimes easier to read. Taiwan and GMSC are the most distinct cases, which we treat as nonlinear regimes because the flexible models improve substantially over logistic regression. At $\lambda=0$, the neural model gives AUC gains of about $5.5$ and $18.7$ percentage points, respectively. \texttt{findr} recovers most of this gain, reaching AUC $0.774$ in Taiwan and $0.850$ in GMSC. The Brier scores follow the same pattern, with the flexible models giving lower prediction error than the linear logistic case. In the other six datasets, the three models are much closer at $\lambda=0$, suggesting approximately linear regimes where the linear baseline already captures most of the predictive signal. This interpretation is further supported by the structured-component diagnostics in Subsection~\ref{subsec:app_inter}.

The full $\lambda$ paths (Figure~\ref{fig:app-performance}) add context beyond the unpenalised cut. Taiwan and Vehicle Loan show the largest predictive cost at high penalties. At $\lambda=1$, AUC falls from $0.774$ to $0.665$ for \texttt{findr} in Taiwan and from $0.641$ to $0.513$ in Vehicle Loan. The neural and logistic paths show similar deterioration in Vehicle Loan, where the high-penalty solutions approach nearly uninformative discrimination. GMSC behaves differently. The flexible models lose AUC as $\lambda$ increases, but they remain well above the logistic baseline through moderate penalties. This indicates that the nonlinear signal remains useful even after adding the fairness penalty.

The remaining datasets show milder or more metric-specific changes. South German and UK Credit retain relatively stable AUC across the path, although South German shows a visible increase in Brier score at larger penalties. Thomas shows modest Brier changes but lower AUC under stronger penalisation. PAKDD 2010 is the main exception to a simple performance-loss description. Its AUC declines as $\lambda$ increases, but its Brier score improves for all three models, suggesting that the penalty reduces discrimination while producing less extreme probability forecasts. In summary, Figure~\ref{fig:app-performance} shows that \texttt{findr} follows the neural model when nonlinear predictive structure is useful and stays close to logistic regression in approximately linear regimes. 

Although not the focus of this paper, we note that Figure~\ref{fig:app-performance} also supports the focus on (semi-)parametric model structures. The unconstrained \texttt{findr}, \texttt{LR}, and \texttt{NN} models typically perform close to CatBoost, indicating that alternative model classes (e.g., tree-based ensemble models) would not provide substantially higher predictive performance.

\begin{table}[!t]
  \centering
  \caption{Out-of-sample AUC and Brier score for the unpenalised models. Bold marks the best result per metric within each row. Datasets are grouped by predictive regime.}
  \label{tab:summary_lam0}
  \setlength{\tabcolsep}{5pt}
\begin{tabular}{lcccccc}
\toprule
 & \multicolumn{6}{c}{$\lambda = 0$} \\
\cmidrule(lr){2-7}
 & \multicolumn{3}{c}{AUC\;($\uparrow$)} & \multicolumn{3}{c}{Brier\;($\downarrow$)} \\
\cmidrule(lr){2-4}\cmidrule(lr){5-7}
Dataset & LR & NN & \texttt{findr} & LR & NN & \texttt{findr} \\
\midrule
\multicolumn{7}{l}{\textit{Nonlinear regime}} \\[2pt]
Taiwan & 0.722 & \textbf{0.777} (0.001) & 0.774 (0.002) & 0.144 & \textbf{0.134} (0.000) & 0.135 (0.000) \\
GMSC & 0.669 & \textbf{0.856} (0.002) & 0.850 (0.003) & 0.059 & \textbf{0.050} (0.000) & 0.052 (0.000) \\
\midrule
\multicolumn{7}{l}{\textit{Linear regime}} \\[2pt]
South German & 0.818 & 0.817 (0.003) & \textbf{0.823} (0.008) & 0.147 & 0.149 (0.001) & \textbf{0.147} (0.003) \\
Vehicle Loan & 0.631 & 0.638 (0.001) & \textbf{0.641} (0.001) & 0.164 & 0.163 (0.000) & \textbf{0.162} (0.000) \\
MyHome & 0.615 & 0.610 (0.003) & \textbf{0.616} (0.002) & 0.232 & 0.232 (0.001) & \textbf{0.231} (0.000) \\
Thomas & 0.589 & \textbf{0.603} (0.006) & 0.597 (0.007) & 0.201 & 0.201 (0.001) & \textbf{0.200} (0.001) \\
UK Credit & 0.712 & \textbf{0.729} (0.002) & 0.722 (0.005) & 0.039 & \textbf{0.039} (0.000) & 0.039 (0.000) \\
PAKDD 2010 & 0.582 & 0.585 (0.001) & \textbf{0.587} (0.001) & 0.210 & \textbf{0.208} (0.000) & 0.211 (0.000) \\
\bottomrule
\end{tabular}

\end{table}

\begin{table}[!t]
  \centering
  \caption{Out-of-sample AUC and Brier score for the three models at $\lambda=0.1$. Bold marks the best result per metric within each row. Datasets are grouped by predictive regime.}
  \label{tab:summary_lam01}
  \setlength{\tabcolsep}{5pt}
\begin{tabular}{lcccccc}
\toprule
 & \multicolumn{6}{c}{$\lambda = 0.1$} \\
\cmidrule(lr){2-7}
 & \multicolumn{3}{c}{AUC\;($\uparrow$)} & \multicolumn{3}{c}{Brier\;($\downarrow$)} \\
\cmidrule(lr){2-4}\cmidrule(lr){5-7}
Dataset & LR & NN & \texttt{findr} & LR & NN & \texttt{findr} \\
\midrule
\multicolumn{7}{l}{\textit{Nonlinear regime}} \\[2pt]
Taiwan & 0.718 & 0.772 (0.002) & \textbf{0.774} (0.001) & 0.147 & 0.136 (0.001) & \textbf{0.135} (0.000) \\
GMSC & 0.669 & 0.833 (0.019) & \textbf{0.849} (0.003) & 0.059 & \textbf{0.051} (0.001) & 0.052 (0.000) \\
\midrule
\multicolumn{7}{l}{\textit{Linear regime}} \\[2pt]
South German & 0.819 & 0.822 (0.004) & \textbf{0.824} (0.004) & 0.151 & 0.148 (0.002) & \textbf{0.147} (0.002) \\
Vehicle Loan & 0.595 & 0.500 (0.000) & \textbf{0.611} (0.011) & 0.169 & 0.170 (0.000) & \textbf{0.169} (0.000) \\
MyHome & \textbf{0.614} & 0.610 (0.003) & 0.611 (0.003) & 0.234 & \textbf{0.233} (0.000) & 0.233 (0.000) \\
Thomas & 0.589 & \textbf{0.604} (0.007) & 0.600 (0.010) & 0.199 & \textbf{0.198} (0.001) & 0.198 (0.001) \\
UK Credit & 0.709 & \textbf{0.726} (0.002) & 0.725 (0.003) & 0.039 & 0.039 (0.000) & \textbf{0.039} (0.000) \\
PAKDD 2010 & 0.583 & 0.584 (0.001) & \textbf{0.587} (0.001) & 0.202 & \textbf{0.202} (0.000) & 0.203 (0.000) \\
\bottomrule
\end{tabular}

\end{table}

\subsubsection{Fairness}
\begin{figure}[!t]
  \centering
  \includegraphics[width=0.9\linewidth]{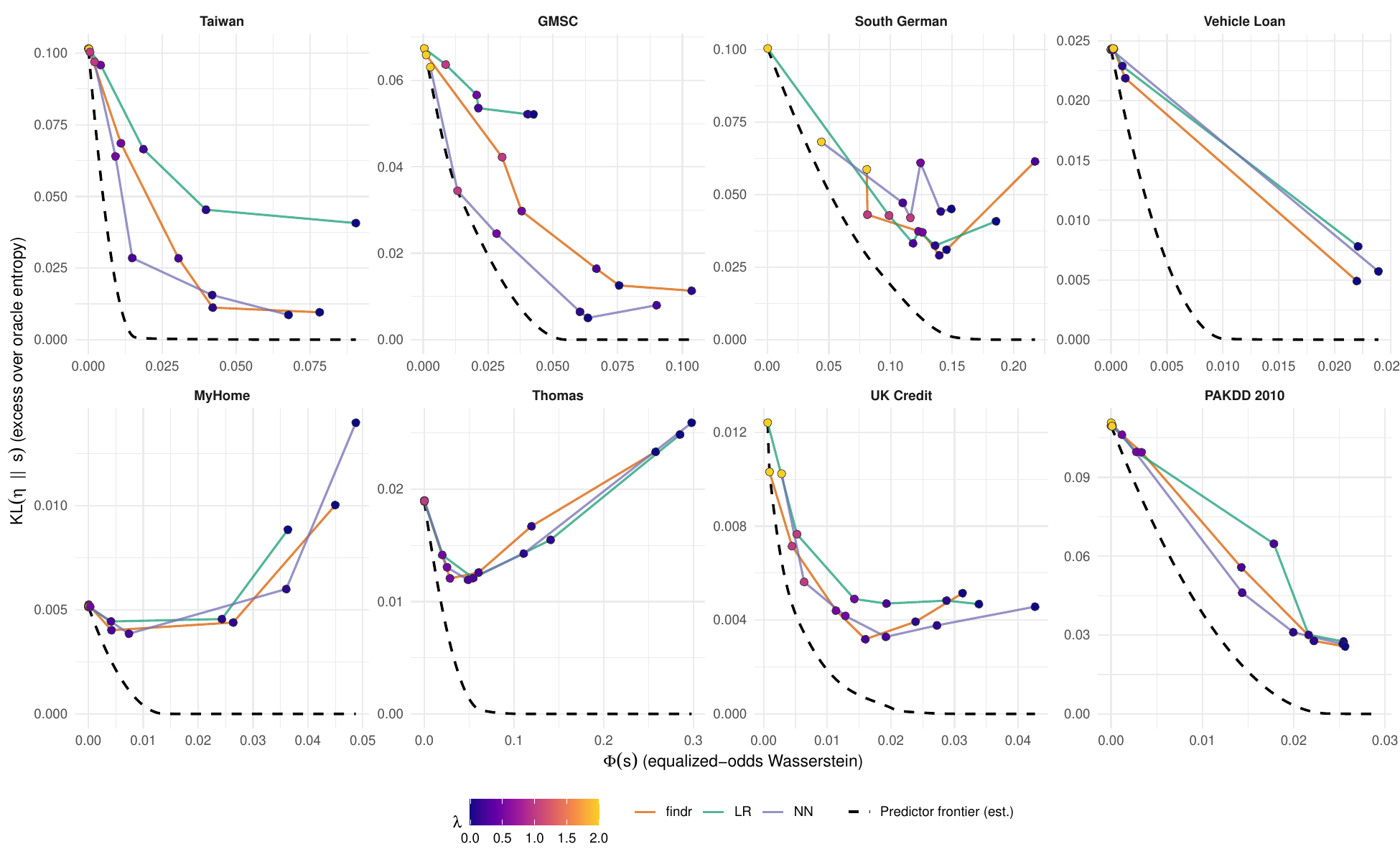}
  \caption{
Score-level predictor frontiers and fitted model paths for the application datasets under equalised odds. The horizontal axis is the empirical equalised-odds Wasserstein gap, and the vertical axis is excess expected log loss relative to the reference risk. The colour of each point indicates the value of $\lambda$.
}
  \label{fig:app-frontier}
\end{figure}

Each panel in Figure~\ref{fig:app-frontier} corresponds to a dataset, showing the score-level predictor frontiers and model paths as $\lambda$ varies.

The frontier comparison shows how each model moves through the score-level accuracy--fairness trade-off as the penalty increases. The main pattern is heterogeneous across datasets. In the nonlinear regimes, \texttt{findr} tends to remain close to the flexible model and closer to the frontier than logistic regression. In the approximately linear regimes, its path is usually comparable to the simpler alternatives and is not systematically worse. The main exception is South German, where the logistic path reaches the low-gap frontier region more directly than \texttt{findr} and \texttt{NN}.

The individual panels support this pattern in different ways. Thomas and South German have the widest fairness scales, but they do not show the same trade-off. In Thomas, all three models reduce the fairness gap with little visible increase in excess loss. South German is more model-dependent, with logistic regression moving more directly toward the low-gap region. These two datasets are also the smallest datasets (see Appendix~\ref{app:data}), which may contribute to the wider and less stable fitted paths because fairness distance is estimated within outcome-by-group cells. 

Taiwan and GMSC are the nonlinear predictive regimes identified in Subsection~\ref{subsec:app_perf}. In both panels, the flexible paths are closer to the estimated frontier than logistic regression over the whole path. This is most visible in GMSC, where \texttt{NN} and \texttt{findr} begin with much lower excess loss than logistic regression at moderate fairness gaps. In Taiwan, all models can drive the fairness gap close to zero, but with a large increase in excess loss. For \texttt{findr}, the gap falls from approximately $0.08$ at $\lambda=0$ to nearly zero at $\lambda=2$, while excess loss rises from about $0.01$ to $0.10$. GMSC has a similar leftward movement, but the flexible paths retain a clearer advantage at lower and moderate penalties. This is consistent with the performance results showing that nonlinear signal is useful in these two datasets.

Vehicle Loan, UK Credit, MyHome, and PAKDD 2010 have smaller fairness scales than Thomas, South German, Taiwan, and GMSC. Vehicle Loan and UK Credit show near-zero gaps at high $\lambda$ with moderate vertical movement. MyHome is also compact, with gaps below about $0.05$, and all three model paths move close to the low-gap region. PAKDD 2010 shows a different pattern. Its initial gaps are already small, near $0.025$, but forcing them close to zero produces one of the largest excess-loss increases in the figure. This cautions against reading all predictive summaries in the same way. The frontier's log-loss scale shows a cost, whereas Tables~\ref{tab:summary_lam0} and~\ref{tab:summary_lam01} show improved Brier scores at the reported $\lambda$ values.

\subsubsection{Interpretability} \label{subsec:app_inter}
\begin{figure}[!t]
  \centering
  \includegraphics[width=0.9\linewidth]{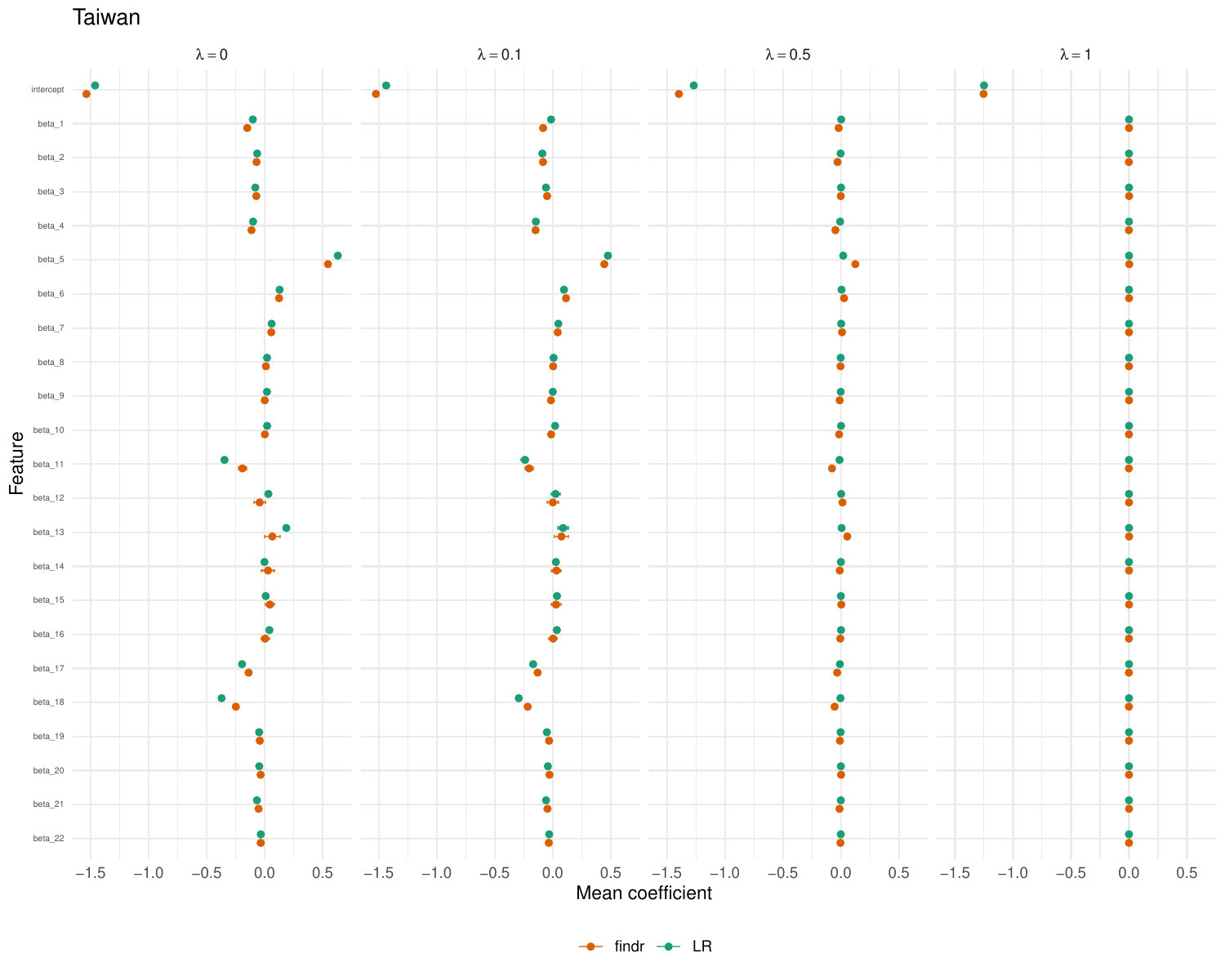}
  \caption{Taiwan Credit Default structured coefficients across seeds under equalised odds for selected values of $\lambda$.}
  \label{fig:coef_taiwan}
\end{figure}

\begin{figure}[!t]
  \centering
  \includegraphics[width=0.9\linewidth]{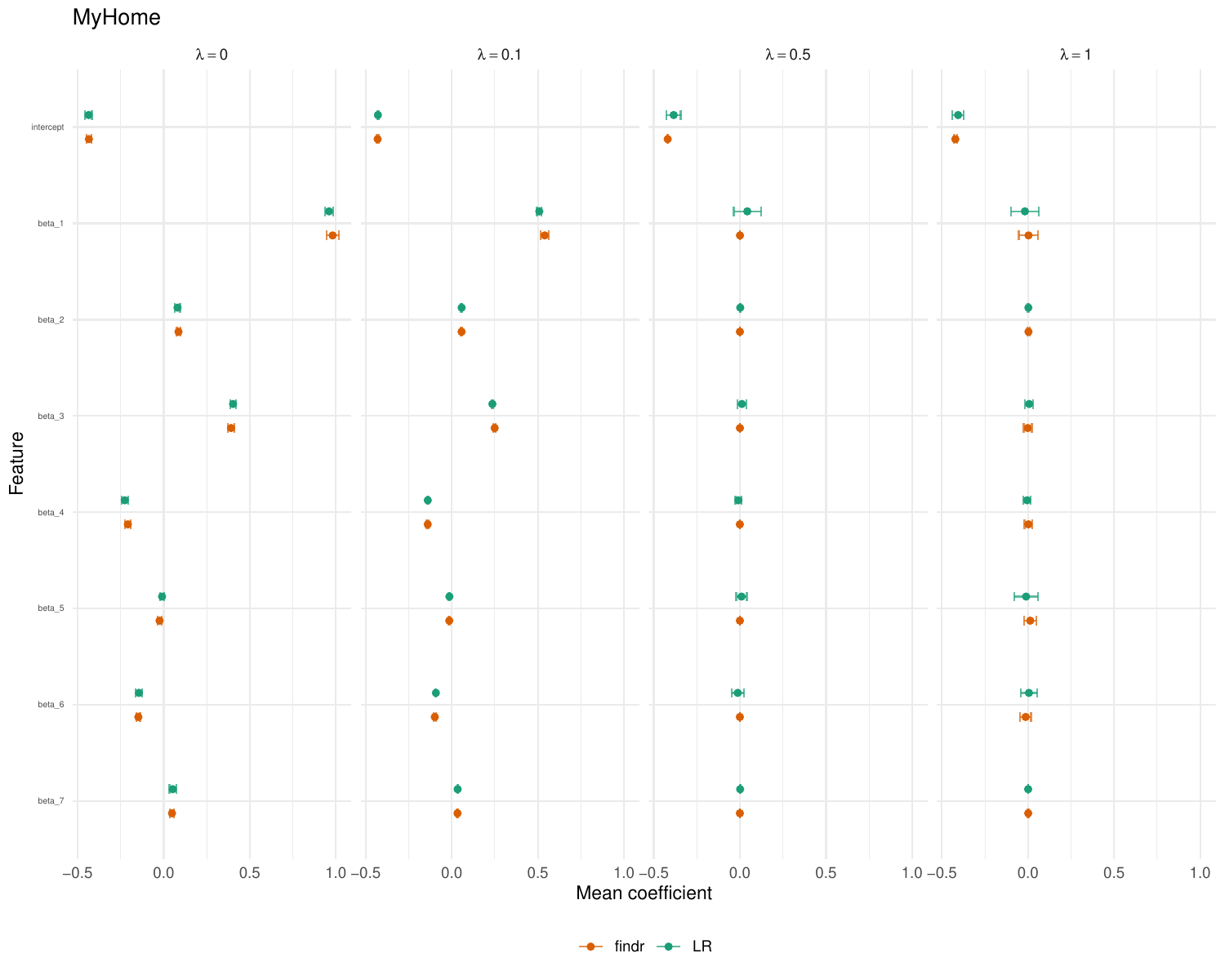}
  \caption{MyHome Loan structured coefficients across seeds under equalised odds for selected values of $\lambda$.}
  \label{fig:coef_myhome}
\end{figure}

The structured part of \texttt{findr} gives a coefficient vector that can be read in the same way as in logistic regression. Figures~\ref{fig:coef_taiwan} and~\ref{fig:coef_myhome} show this information for Taiwan and MyHome, which represent a nonlinear and an approximately linear regime, respectively. Analogous coefficient plots are obtained for the other datasets. Since the coefficients are displayed in generic feature notation, we focus on their sign, magnitude, and stability rather than on feature-specific substantive interpretation. In both examples, the structured coefficients remain readable across the selected values of $\lambda$, even though the full model also includes a neural residual.

We are also interested in whether these coefficients are enough to explain the fitted decisions. As in the simulation study, a readable coefficient vector does not imply that the linear component describes the whole fitted risk surface. Tables~\ref{tab:summary_evr_ddr_cser_lam0} and~\ref{tab:summary_evr_ddr_cser_lam01} report this comparison for $\lambda=0$ and $\lambda=0.1$. Recall that EVR measures how much logit-scale variation is assigned to the structured component, DDR measures decision agreement between the structured component and the full model, and CSER measures whether local coefficient directions remain consistent after adding the residual.

\begin{table}[!t]
  \centering
  \caption{EVR, DDR, and CSER for \texttt{findr} at $\lambda=0$ under equalised odds. Entries are mean and standard deviation across seeds. Datasets are grouped by predictive regime.}
  \label{tab:summary_evr_ddr_cser_lam0}
  \setlength{\tabcolsep}{5pt}
\begin{tabular}{lcccc}
\toprule
 & \multicolumn{4}{c}{$\lambda = 0$} \\
\cmidrule(lr){2-5}
Dataset & EVR & DDR & Min CSER & Max CSER \\
\midrule
\multicolumn{5}{l}{\textit{Nonlinear regime}} \\[2pt]
Taiwan & 0.527 (0.008) & 0.084 (0.002) & 0.088 (0.050) & 0.470 (0.046) \\
GMSC & 0.399 (0.034) & 0.034 (0.004) & 0.018 (0.010) & 0.215 (0.195) \\
\midrule
\multicolumn{5}{l}{\textit{Linear regime}} \\[2pt]
South German & 0.957 (0.017) & 0.030 (0.013) & 0.000 (0.000) & 0.158 (0.126) \\
Vehicle Loan & 0.805 (0.014) & 0.004 (0.001) & 0.001 (0.001) & 0.204 (0.017) \\
MyHome & 0.935 (0.024) & 0.022 (0.006) & 0.000 (0.000) & 0.004 (0.004) \\
Thomas & 0.981 (0.015) & 0.005 (0.006) & 0.000 (0.000) & 0.051 (0.071) \\
UK Credit & 0.749 (0.024) & 0.000 (0.000) & 0.051 (0.016) & 0.429 (0.012) \\
PAKDD 2010 & 0.946 (0.008) & 0.028 (0.003) & 0.000 (0.000) & 0.044 (0.032) \\
\bottomrule
\end{tabular}

\end{table}

\begin{table}[!t]
  \centering
  \caption{EVR, DDR, and CSER for \texttt{findr} at $\lambda=0.1$ under equalised odds. Entries are mean and standard deviation across seeds. Datasets are grouped by predictive regime.}
  \label{tab:summary_evr_ddr_cser_lam01}
  \setlength{\tabcolsep}{5pt}
\begin{tabular}{lcccc}
\toprule
 & \multicolumn{4}{c}{$\lambda = 0.1$} \\
\cmidrule(lr){2-5}
Dataset & EVR & DDR & Min CSER & Max CSER \\
\midrule
\multicolumn{5}{l}{\textit{Nonlinear regime}} \\[2pt]
Taiwan & 0.505 (0.009) & 0.095 (0.002) & 0.039 (0.028) & 0.462 (0.034) \\
GMSC & 0.392 (0.019) & 0.021 (0.002) & 0.011 (0.012) & 0.145 (0.040) \\
\midrule
\multicolumn{5}{l}{\textit{Linear regime}} \\[2pt]
South German & 0.965 (0.011) & 0.033 (0.013) & 0.000 (0.000) & 0.147 (0.085) \\
Vehicle Loan & 0.916 (0.020) & 0.000 (0.000) & 0.000 (0.000) & 0.193 (0.124) \\
MyHome & 0.876 (0.036) & 0.008 (0.004) & 0.007 (0.021) & 0.067 (0.054) \\
Thomas & 0.952 (0.033) & 0.002 (0.003) & 0.000 (0.000) & 0.093 (0.101) \\
UK Credit & 0.753 (0.014) & 0.000 (0.000) & 0.016 (0.011) & 0.401 (0.008) \\
PAKDD 2010 & 0.927 (0.005) & 0.022 (0.002) & 0.000 (0.000) & 0.120 (0.058) \\
\bottomrule
\end{tabular}

\end{table}

In the approximately linear regime datasets, the coefficients are close to sufficient for describing the fitted decisions. At $\lambda=0$, EVR is $0.957$ in South German, $0.935$ in MyHome, $0.981$ in Thomas, and $0.946$ in PAKDD 2010. DDR is also small in these datasets, so the structured component usually gives the same binary decision as the full model. These are the cases where the interpretation is closest to logistic regression, with the additional check that little decision-relevant information is left to the residual. Vehicle Loan and UK Credit are more intermediate. Their EVR values are lower, $0.805$ and $0.749$, but DDR remains close to zero. The residual therefore contributes to continuous score variation, with limited effect on thresholded decisions.

Taiwan and GMSC show why having coefficients is not the same as having a complete coefficient-based explanation. Their EVR values are much lower, $0.527$ and $0.399$ at $\lambda=0$ and $0.505$ and $0.392$ at $\lambda=0.1$, which means that the neural residual carries a substantial share of logit-scale variation. In Taiwan, nonzero DDR and high maximum CSER suggest that the residual can affect both some decisions and some local directional interpretations. In GMSC, DDR remains low despite low EVR, so the residual changes the continuous score surface more than the final thresholded decisions. 

Read together with the coefficient plots, these results clarify the role of the structured component. In MyHome, the plotted coefficients are close to a full decision-level explanation. In Taiwan, they remain meaningful, but the residual must be examined to understand the full fitted model. In this way, \texttt{findr} keeps the familiar coefficient interpretation when it is adequate, while the proposed diagnostics indicate when additional nonlinear information must also be considered.

\section{Conclusion}

In high-stakes decisions such as credit scoring, there is a common tension between performance, interpretability, and fairness. Linear models provide coefficients that are easy to interpret and audit, but they can miss nonlinear structure or higher-order interactions. Flexible models can accommodate these predictive effects, but they usually do not by themselves identify which part of the fitted score is interpretable, nor whether structured explanations are sufficient for decisions.

We introduce \texttt{findr}, a semi-structured approach for binary regression that combines a readable linear logit, an orthogonal neural residual, and an in-processing Wasserstein fairness penalty. The model is interpretable through its coefficient component, flexible through its neural residual, and fairness-aware through a penalty that compares score distributions during training. The additive decomposition is useful beyond the specific linear specification used in the empirical analysis. It can identify a structured and interpretable component, which could be replaced by more expressive structured models such as GAMs, while keeping the residual component available for information that is difficult to represent in a standard structured model. This also opens the possibility of incorporating richer data sources, such as text, through the neural residual. The flexibility of the fairness penalisation allows us to use the same construction with different fairness notions by changing which conditional score distributions enter the Wasserstein distance.

We also contribute tools for evaluating and comparing the fitted models. The interpretability diagnostics separate three questions. EVR measures how much logit-scale variation is assigned to the structured component. DDR measures whether the structured and full logits lead to the same binary decisions. CSER identifies cases where the direction implied by a specific coefficient differs from the full model's local response. The score-level frontiers provide an empirical benchmark for how much predictive loss is associated with lower fairness gaps on the same validation sample.

We illustrate the approach in a simulation study. When the true signal is linear, \texttt{findr} behaves like logistic regression, with most fitted variation assigned to the structured component and little disagreement between the structured and full decisions. When the signal contains nonlinearities or interactions, the residual carries a larger share of the logit-scale variation, while the structured coefficients remain coherent with the linear part of the signal. The Wasserstein penalty moves scores toward lower fairness gaps by acting on entire score distributions rather than on a single threshold or a single distributional summary.

We apply the approach to eight credit-related datasets. The results are more heterogeneous, as expected in real applications. In Taiwan and GMSC, the nonlinear predictive component is relevant, and \texttt{findr} recovers much of the gain of the neural model while retaining a structured coefficient component. In other datasets, the linear baseline already captures most of the predictive signal. We also show that fairness costs vary across datasets and across evaluation criteria. Some datasets allow large reductions in the Wasserstein gap with little excess log-loss cost, while others require stronger changes to the fitted scores. We then use the interpretability diagnostics to clarify when the coefficient component is close to a full decision-level explanation and when the residual must also be examined.

Overall, our results support semi-structured modelling as a practical middle ground between logistic regression and fully flexible neural models. We do not remove the need to choose a fairness criterion or to evaluate model behaviour in context. We do, however, make the performance, fairness, and interpretability trade-offs more explicit. In settings where the linear explanation is adequate, \texttt{findr} behaves close to the transparent baseline. In settings where nonlinear structure matters, it can use that structure while reporting how much decision-relevant information remains outside the coefficient component.

Several extensions follow from this framework. On the fairness side, future work could extend the Wasserstein penalties to predefined multi-group settings, intersectional group definitions, and individual-based notions such as counterfactual fairness. On the interpretability side, the diagnostics suggest a route toward variable selection under an explicit notion of interpretability, where the goal is to retain the variables that carry most of the interpretable signal while leaving remaining predictive information to the residual.

\bibliographystyle{apalike}  
\bibliography{references} 

\appendix

\section{Computational construction of the predictor frontier}
\label{app:predictor-frontier}

This appendix summarises the numerical construction used for the score-level frontier in Section~\ref{sec:frontier}. For a score vector $s$, the code evaluates $\widehat\Phi(s;y,a)$ by exact step-function integration of the empirical CDF gaps within the realised validation cells $(Y=y,A=a)$. It evaluates $\widehat R_{\log}(r,s)$ by passing $r$ as the first argument to the log-loss routine, so the loss is the expected Bernoulli log loss under the reference risk rather than the realised validation log loss.

For each frontier multiplier $\xi$, the scalarised objective is
\[
Q_\xi(s)=\widehat R_{\log}(r,s)+\xi\widehat\Phi(s;y,a).
\]
The default grid is $\{0\}\cup \operatorname{logspace}(-2,2,40)$. The case $\xi=0$ is returned in closed form as $s=r$. For $\xi>0$, the solver starts from the candidate in the pool consisting of $r$, optional fitted-model score vectors, and the previous multiplier solution that gives the smallest value of $Q_\xi$.

The iteration uses a soft-cell approximation to the fairness direction. For cells $(u,g)$, define
\[
\pi_{ug}(x_i)=\mathbf 1(a_i=g)\{r_i\mathbf 1(u=1)+(1-r_i)\mathbf 1(u=0)\},
\qquad
p_{ug}=n^{-1}\sum_i \pi_{ug}(x_i),
\]
and let $F_{ug}(v;s)$ be the corresponding weighted empirical CDF of the current scores. The effective sign used in the update is
\begin{equation}
    \Theta_i
    = \sum_{u\in\{0,1\}}\sum_{g\in\{0,1\}}
      \frac{\pi_{ug}(x_i)}{p_{ug}}
      \operatorname{sign}\{F_{ug}(s_i;s)-F_{u,1-g}(s_i;s)\}.
    \label{eq:theta}
\end{equation}
The score update is
\begin{equation}
    s_i^{(t+1)}=(1-\rho_\xi)s_i^{(t)}
    +\rho_\xi\left\{r_i+\xi s_i^{(t)}(1-s_i^{(t)})\Theta_i^{(t)}\right\},
    \qquad
    \rho_\xi=\frac{\rho}{1+\xi},
    \label{eq:update}
\end{equation}
followed by clipping to $[\varepsilon,1-\varepsilon]$. A positive $\Theta_i$ increases the next score for observation $i$. Since larger scores mean higher predicted default risk, this direction raises the default probability assigned to observations whose soft outcome-cell memberships place them in cells with relatively larger weighted CDF mass at their current score. It moves mass to the right in those cells. The objective used to accept iterates is still the realised-cell objective $Q_\xi$, so this update should be read as a numerical heuristic rather than an exact subgradient step for $\widehat\Phi(s;y,a)$.

The solver retains the best objective value among the initial score, periodic iterates, and the Polyak--Ruppert average of the second half of the path. Candidate scores are then added to the point set before envelope extraction, because they are feasible score vectors on the same validation sample. If enabled, one refinement pass inserts up to fifteen additional multipliers where adjacent solver-produced points have large normalised chord lengths. The current implementation chooses these new multipliers by log-midpoints, with the special case $\xi_0=0<\xi_1$ using $\xi_1/2$. The final frontier sorts all points by $\widehat\Phi$ and takes the running minimum of $\widehat R_{\log}$, collapsing duplicate unfairness values.

\section{Datasets summary statistics} \label{app:data}

\begin{table}[ht]
\centering
\caption{Dataset summary statistics. Default rates are reported overall and by sensitive group ($A=0$ vs.\ $A=1$).}
\label{tab:dataset_summary}
\begin{tabular}{lrrrrr}
\toprule
Dataset & $n$ & Default rate (\%) & Sensitive share (\%) & Default rate $A{=}0$ (\%) & Default rate $A{=}1$ (\%) \\
\midrule
South German  & 1{,}000   & 30.00 & 19.00 & 27.16 & 42.11 \\
Taiwan        & 30{,}000  & 22.12 & 12.90 & 21.45 & 26.66 \\
GMSC          & 150{,}000 & 6.68  & 2.02  & 6.59  & 11.16 \\
Vehicle loan  & 233{,}154 & 21.71 & 22.84 & 21.03 & 24.00 \\
MyHome        & 7{,}000   & 40.00 & 7.03  & 39.67 & 44.31 \\
Thomas        & 1{,}225   & 26.37 & 7.67  & 24.49 & 48.94 \\
UK            & 30{,}000  & 4.00  & 19.75 & 3.55  & 5.84  \\
PAKDD         & 50{,}000  & 26.08 & 11.49 & 25.02 & 34.29 \\
\bottomrule
\end{tabular}
\end{table}

\section{Hyperparameter search}
\label{app:hyperparameter-search}

Table~\ref{tab:application-hyperparameters} reports the hyperparameter space used for the empirical experiments. We use discrete random search separately for each value of $\lambda$. Candidate configurations are ranked by their mean validation objective over two optimisation restarts. Training is limited to 6,000 epochs, with early stopping after 20 epochs without an improvement during tuning. All models are trained with AdamW and binary cross-entropy with logits. After selecting a configuration, each experiment is repeated over 10 random seeds and evaluated on the held-out test set.

\begin{table}[ht]
  \centering
  \caption{Hyperparameter search space.}
  \label{tab:application-hyperparameters}
  \small
  \begin{tabular}{@{}ll@{}}
    \toprule
    Hyperparameter & Candidate values \\
    \midrule
    Learning rate
      & \(\{3\!\times\!10^{-4},10^{-3},3\!\times\!10^{-3},10^{-2}\}\) \\
    Weight decay
      & \(\{0,10^{-5},10^{-4},10^{-3}\}\) \\
    Batch size
      & \(\{32,64,128,256\}\) \\
    Hidden-layer widths
      & \(\{[4],[16,8],[64,32],[128,64]\}\) \\
    Activation
      & \{ReLU, GELU, tanh\} \\
    Dropout rate
      & \(\{0,0.1,0.25\}\) \\
    \bottomrule
  \end{tabular}
\end{table}

\end{document}